# From Predictive to Prescriptive Analytics

## Dimitris Bertsimas

Sloan School of Management, Massachusetts Institute of Technology, Cambridge, MA 02139, dbertsim@mit.edu

## Nathan Kallus

School of Operations Research and Information Engineering and Cornell Tech, Cornell University, New York, NY 11109, kallus@cornell.edu

In this paper, we combine ideas from machine learning (ML) and operations research and management science (OR/MS) in developing a framework, along with specific methods, for using data to prescribe optimal decisions in OR/MS problems. In a departure from other work on data-driven optimization and reflecting our practical experience with the data available in applications of OR/MS, we consider data consisting, not only of observations of quantities with direct effect on costs/revenues, such as demand or returns, but predominantly of observations of associated auxiliary quantities. The main problem of interest is a conditional stochastic optimization problem, given imperfect observations, where the joint probability distributions that specify the problem are unknown. We demonstrate that our proposed solution methods, which are inspired by ML methods such as local regression (LOESS), classification and regression trees (CART), and random forests (RF), are generally applicable to a wide range of decision problems. We prove that they are computationally tractable and asymptotically optimal under mild conditions even when data is not independent and identically distributed (iid) and even for censored observations. We extend these results to the case where some of the decision variables can directly affect uncertainty in unknown ways, such as pricing's effect on demand in joint pricing and planning problems. As an analogue to the coefficient of determination $R^2$, we develop a metric $P$ termed the coefficient of prescriptiveness to measure the prescriptive content of data and the efficacy of a policy from an operations perspective. To demonstrate the power of our approach in a real-world setting we study an inventory management problem faced by the distribution arm of an international media conglomerate, which ships an average of 1 billion units per year. We leverage both internal data and public online data harvested from IMDb, Rotten Tomatoes, and Google to prescribe operational decisions that outperform baseline measures. Specifically, the data we collect, leveraged by our methods, accounts for an 88% improvement as measured by our coefficient of prescriptiveness.

## 1. Introduction

In today's data-rich world, many problems of operations research and management science (OR/MS) can be characterized by three primitives:

a) Data $\{y^1, \ldots, y^N\}$ on uncertain quantities of interest $Y \in \mathcal{Y} \subset \mathbb{R}^{d_y}$ such as simultaneous demands.

b) Auxiliary data $\{x^1, \ldots, x^N\}$ on associated covariates $X \in \mathcal{X} \subset \mathbb{R}^{d_x}$ such as recent sale figures, volume of Google searches for a products or company, news coverage, or user reviews, where $x^i$ is concurrently observed with $y^i$.

c) A decision $z$ constrained in $\mathcal{Z} \subset \mathbb{R}^{d_z}$ made after some observation $X = x$ with the objective of minimizing the *uncertain* costs $c(z; Y)$.

Traditionally, decision-making under uncertainty in OR/MS has largely focused on the problem

$$v^{\text{stoch}} = \min_{z \in \mathcal{Z}} \mathbb{E}\left[c(z; Y)\right], \qquad z^{\text{stoch}} \in \arg\min_{z \in \mathcal{Z}} \mathbb{E}\left[c(z; Y)\right] \qquad (1)$$

and its multi-period generalizations and addressed its solution under a priori assumptions about the distribution $\mu_Y$ of $Y$ (cf. Birge and Louveaux (2011)), or, at times, in the presence of data $\{y^1, \ldots, y^n\}$ in the assumed form of independent and identically distributed (iid) observations drawn from $\mu_Y$ (cf. Shapiro (2003), Shapiro and Nemirovski (2005), Kleywegt et al. (2002)). (We will discuss examples of (1) in Section 1.1.) By and large, auxiliary data $\{x^1, \ldots, x^N\}$ has not been extensively incorporated into OR/MS modeling, despite its growing influence in practice.

From its foundation, machine learning (ML), on the other hand, has largely focused on supervised learning, or the prediction of a quantity $Y$ (usually univariate) as a function of $X$, based on data $\{(x^1, y^1), \ldots, (x^N, y^N)\}$. By and large, ML does not address optimal decision-making under uncertainty that is appropriate for OR/MS problems.

At the same time, an explosion in the availability and accessibility of data and advances in ML have enabled applications that predict, for example, consumer demand for video games ($Y$) based on online web-search queries ($X$) (Choi and Varian (2012)) or box-office ticket sales ($Y$) based on Twitter chatter ($X$) (Asur and Huberman (2010)). There are many other applications of ML that proceed in a similar manner: use large-scale auxiliary data to generate predictions of a quantity that is of interest to OR/MS applications (Goel et al. (2010), Da et al. (2011), Gruhl et al. (2005, 2004), Kallus (2014)). However, it is not clear how to go from a good prediction to a good decision. A good decision must take into account uncertainty wherever present. For example, in the absence of auxiliary data, solving (1) based on data $\{y^1, \ldots, y^n\}$ but using only the sample mean $\overline{y} = \frac{1}{N}\sum_{i=1}^{N} y^i \approx \mathbb{E}\left[Y\right]$ and ignoring all other aspects of the data would generally lead to inadequate solutions to (1) and an unacceptable waste of good data.

In this paper, we combine ideas from ML and OR/MS in developing a framework, along with specific methods, for using data to prescribe optimal decisions in OR/MS problems that leverage auxiliary observations. Specifically, the problem of interest is

$$v^*(x) = \min_{z \in \mathcal{Z}} \mathbb{E}\left[c(z; Y) \big| X = x\right], \qquad z^*(x) \in \mathcal{Z}^*(x) = \arg\min_{z \in \mathcal{Z}} \mathbb{E}\left[c(z; Y) \big| X = x\right], \qquad (2)$$

where the underlying distributions are unknown and only data $S_N = \{(x^1, y^1), \ldots, (x^N, y^N)\}$ is available. The solution $z^*(x)$ to (2) represents the full-information optimal decision, which, via full knowledge of the unknown joint distribution $\mu_{X,Y}$ of $(X, Y)$, leverages the observation $X = x$ to the fullest possible extent in minimizing costs. We use the term *predictive prescription* for any function $z(x)$ that prescribes a decision in anticipation of the future given the observation $X = x$. Our task is to use $S_N$ to construct a data-driven predictive prescription $\hat{z}_N(x)$. Our aim is that its performance in practice, $\mathbb{E}\left[c(\hat{z}_N(x); Y) \middle| X = x\right]$, is close to the full-information optimum, $v^*(x)$.

Our key contributions include:

a) We propose various ways for constructing predictive prescriptions $\hat{z}_N(x)$ The focus of the paper is predictive prescriptions that have the form

$$\hat{z}_N(x) \in \arg\min_{z \in \mathcal{Z}} \sum_{i=1}^N w_{N,i}(x)c(z; y^i), \tag{3}$$

where $w_{N,i}(x)$ are weight functions derived from the data. We motivate specific constructions inspired by a great variety of predictive ML methods, including for example and random forests (RF; Breiman (2001)). We briefly summarize a selection of these constructions that we find the most effective below.

b) We also consider a construction motivated by the traditional empirical risk minimization (ERM) approach to ML. This construction has the form

$$\hat{z}_N(\cdot) \in \arg\min_{z(\cdot) \in \mathcal{F}} \frac{1}{N} \sum_{i=1}^N c(z(x^i); y^i), \tag{4}$$

where $\mathcal{F}$ is some class of functions. We extend the standard ML theory of out-of-sample guarantees for ERM to the case of multivariate-valued decisions encountered in OR/MS problems. We find, however, that in the specific context of OR/MS problems, the construction (4) suffers from some limitations that do not plague the predictive prescriptions derived from (3).

c) We show that that our proposals are computationally tractable under mild conditions.

d) We study the asymptotics of our proposals under sampling assumptions more general than iid by leveraging universal law-of-large-number results of Walk (2010). Under appropriate conditions and for certain predictive prescriptions $\hat{z}_N(x)$ we show that costs with respect to the true distributions converge to the full information optimum, i.e., $\lim_{N \to \infty} \mathbb{E}\left[c(\hat{z}_N(x); Y) \middle| X = x\right] = v^*(x)$, and that prescriptions converge to true full information optimizers, i.e., $\lim_{N \to \infty} \inf_{z \in Z^*(x)} ||z - \hat{z}_N(x)|| = 0$, both for almost everywhere $x$ and almost surely. We extend our results to the case of censored data (such as observing demand via sales).

e) We extend the above results to the case where some of the decision variables may affect the uncertain variable in unknown ways not encapsulated in the known cost function. In this case, the uncertain variable $Y(z)$ will be different depending on the decision and the problem of interest becomes $\min_{z \in \mathcal{Z}} \mathbb{E}\left[c(z; Y(z)) \middle| X = x\right]$. Complicating the construction of a data-driven predictive prescription, however, is that the data only includes the realizations $Y_i = Y_i(Z_i)$ corresponding to historic decisions. For example, in problems that involve pricing decisions such as simultaneous planning and pricing, price has an unknown causal effect on demand that must be determined in order to optimize the full decision $z$, which includes prices and production or shipment plans, and the data only includes demand realized at particular historical prices. We show that under certain conditions our methods can be extended to this case while perserving favorable asymptotic properties.

f) We introduce a new metric $P$, termed *the coefficient of prescriptiveness*, in order to measure the efficacy of a predictive prescription and to assess the prescriptive content of covariates $X$, that is, the extent to which observing $X$ is helpful in reducing costs. An analogue to the coefficient of determination $R^2$ of predictive analytics, $P$ is a unitless quantity that is (eventually) bounded between 0 (not prescriptive) and 1 (highly prescriptive).

g) We demonstrate in a real-world setting the power of our approach. We study an inventory management problem faced by the distribution arm of an international media conglomerate. This entity manages over 0.5 million unique items at some 50,000 retail locations around the world, with which it has vendor-managed inventory (VMI) and scan-based trading (SBT) agreements. On average it ships about 1 billion units a year. We leverage both internal company data and, in the spirit of the aforementioned ML applications, large-scale public data harvested from online sources, including IMDb, Rotten Tomatoes, and Google Trends. These data combined, leveraged by our approach, lead to large improvements in comparison to baseline measures, in particular accounting for an 88% improvement toward the deterministic perfect-foresight counterpart.

Of our proposed constructions of predictive prescriptions $\hat{z}_N(x)$, the ones that we find to be generally the most broadly and practically effective are the following:

a) Motivated by $k$-nearest-neighbors regression ($k$NN; Altman (1992)),

$$\hat{z}_N^{k\text{NN}}(x) \in \arg\min_{z \in \mathcal{Z}} \sum_{i \in \mathcal{N}_k(x)} c(z; y^i), \tag{5}$$

where $\mathcal{N}_k(x) = \{i = 1, \ldots, N : \sum_{j=1}^N \mathbb{I}\left[||x - x_i|| \geq ||x - x_j||\right] \leq k\}$ is the neighborhood of the $k$ data points that are closest to $x$.

b) Motivated by local linear regression (LOESS; Cleveland and Devlin (1988)),

$$\hat{z}_N^{\text{LOESS*}}(x) \in \arg\min_{z \in \mathcal{Z}} \sum_{i=1}^n k_i(x) \max\{1 - \sum_{j=1}^n k_j(x)(x^j - x)^T \Xi(x)^{-1}(x^i - x), 0\} c(z; y^i), \tag{6}$$

where $\Xi(x) = \sum_{i=1}^n k_i(x)(x^i - x)(x^i - x)^T$, $k_i(x) = (1 - (||x^i - x|| / h_N(x))^3)^3 \mathbb{I}\left[||x^i - x|| \leq h_N(x)\right]$, and $h_N(x) > 0$ is the distance to the $k$-nearest point from $x$. Although this form may seem complicated, it (nearly) corresponds to the simple idea of approximating $\mathbb{E}\left[c(z; Y) \big| X = x\right]$ locally by a linear function in $x$, which we will discuss at greater length in Section 2.

c) Motivated by classification and regression trees (CART; Breiman et al. (1984)),

$$\hat{z}_N^{\text{CART}}(x) \in \arg\min_{z \in \mathcal{Z}} \sum_{i : R(x^i) = R(x)} c(z; y^i), \tag{7}$$

where $R(x)$ is the binning rule implied by a regression tree trained on the data $S_N$ as shown in an example in Figure 1.

**Figure 1** A regression tree is trained on data $\{(x^1, y^1), \ldots, (x^{10}, y^{10})\}$ and partitions the $X$ data into regions defined by the leaves. The $Y$ prediction $\hat{m}(x)$ is $\hat{m}_j$, the average of $Y$ data at the leaf in which $X = x$ ends up. The implicit binning rule is $R(x)$, which maps $x$ to the identity of the leaf in which it ends up.

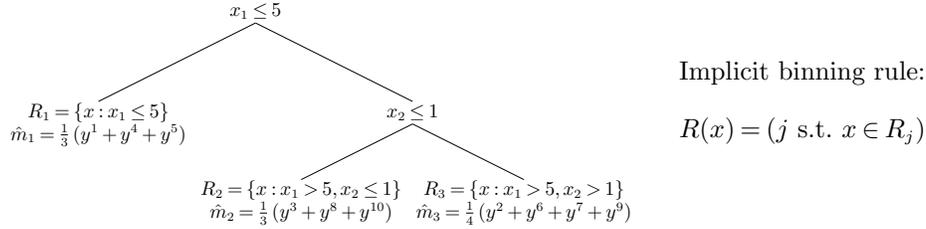

Implicit binning rule:

$$R(x) = (j \text{ s.t. } x \in R_j)$$

d) Motivated by random forests (RF; Breiman (2001)),

$$\hat{z}_N^{\text{RF}}(x) \in \arg\min_{z \in \mathcal{Z}} \sum_{t=1}^{T} \frac{1}{|\{j : R^t(x^j) = R^t(x)\}|} \sum_{i : R^t(x^i) = R^t(x)} c(z; y^i), \tag{8}$$

where where $R^t(x)$ is the binning rule implied by the $t^{\text{th}}$ tree in a random forest trained on the data $S_N$.

Further detail and other constructions are given in Sections 2 and 8.

## 1.1. An Illustrative Examples

In this section, we discuss different approaches to problem (2) and compare them in a two-stage linear decision making problem, illustrating the value of auxiliary data and the methodological gap to be addressed. We illustrate this with synthetic data but, in Section 6, we study a real-world problem and use real-world data.

The specific problem we consider is a two-stage shipment planning problem. We have a network of $d_z$ warehouses that we use in order to satisfy the demand for a product at $d_y$ locations. We consider two stages of the problem. In the first stage, some time in advance, we choose amounts $z_i \geq 0$ of units of product to produce and store at each warehouse $i$, at a cost of $p_1 > 0$ per unit produced. In the second stage, demand $Y \in \mathbb{R}^{d_y}$ realizes at the locations and we must ship units to satisfy it. We can ship from warehouse $i$ to location $j$ at a cost of $c_{ij}$ per unit shipped (recourse variable $s_{ij} \geq 0$) and we have the option of using last-minute production at a cost of $p_2 > p_1$ per unit (recourse variable $t_i$). The overall problem has the cost function and feasible set

$$c(z; y) = p_1 \sum_{i=1}^{d_z} z_i + \min_{(t,s) \in \mathcal{Q}(z,y)} (p_2 \sum_{i=1}^{d_z} t_i + \sum_{i=1}^{d_z} \sum_{j=1}^{d_y} c_{ij} s_{ij}), \quad \mathcal{Z} = \left\{ z \in \mathbb{R}^{d_z} : z \geq 0 \right\},$$

where $\mathcal{Q}(z,y) = \{(s,t) \in \mathbb{R}^{(d_z \times d_y) \times d_z} : t \geq 0, \ s \geq 0, \ \sum_{i=1}^{d_z} s_{ij} \geq y_j \, \forall j, \ \sum_{j=1}^{d_y} s_{ij} \leq z_i + t_i \, \forall i \}$.

The key concern is that we do not know $Y$ or its distribution. We consider the situation where we only have data $S_N = ((x^1, y^1), \ldots, (x^N, y^N))$ consisting of observations of $Y$ along with concurrent observations of some auxiliary quantities $X$ that may be associated with the future value of $Y$. For example, in the portfolio allocation problem, $X$ may include past security returns, behavior of underlying securities, analyst ratings, or volume of Google searches for a company together

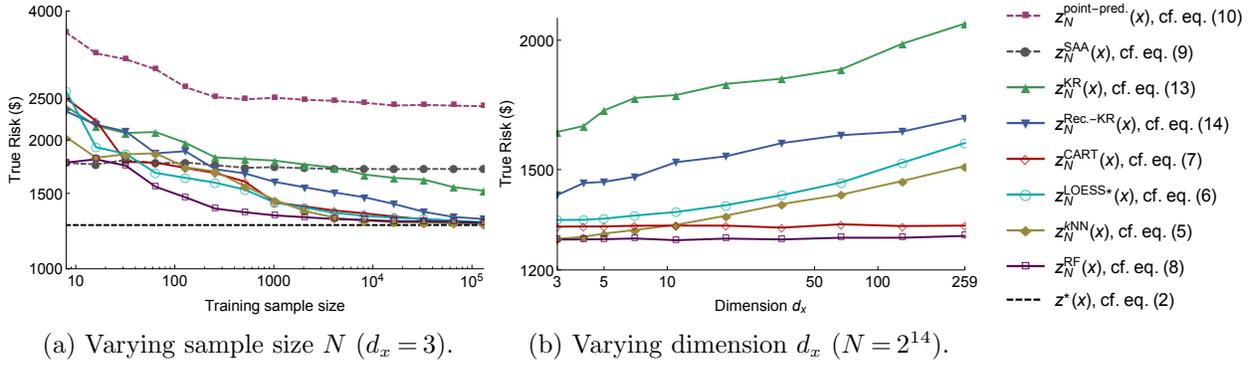

(a) Varying sample size $N$ ($d_x = 3$).   (b) Varying dimension $d_x$ ($N = 2^{14}$).

**Figure 2    Performance of various prescriptions with respect to true distributions, averaged over samples and new observations $x$ (lower is better). Note the horizontal and vertical log scales.**

with keywords like "merger." In the shipment planning problem, $X$ may include, for example, past product sale figures at each of the different retail locations, weather forecasts at the locations, or volume of Google searches for a product to measure consumer attention.

We consider two possible existing data-driven approaches to leveraging such data for making a decision. One approach is the sample average approximation of stochastic optimization (SAA, for short). SAA only concerns itself with the marginal distribution of $Y$, thus ignoring data on $X$, and solves the following data-driven optimization problem

$$\hat{z}_N^{\text{SAA}} \in \arg\min_{z \in \mathcal{Z}} \tfrac{1}{N} \sum_{i=1}^N c(z; y^i), \tag{9}$$

whose objective approximates $\mathbb{E}\left[c(z; Y)\right]$.

Machine learning, on the other hand, leverages the data on $X$ as it tries to predict $Y$ given observations $X = x$. Consider for example a random forest trained on the data $S_N$. It provides a point prediction $\hat{m}_N(x)$ for the value of $Y$ when $X = x$. Given this prediction, one possibility is to consider the approximation of the random variable $Y$ by our best-guess value $\hat{m}_N(x)$ and solve the corresponding optimization problem,

$$\hat{z}_N^{\text{point-pred}} \in \arg\min_{z \in \mathcal{Z}} c(z; \hat{m}_N(x)). \tag{10}$$

The objective approximates $c\left(z; \mathbb{E}\left[Y \big| X = x\right]\right)$. We call (10) a point-prediction-driven decision.

If we knew the full joint distribution of $Y$ and $X$, then the optimal decision having observed $X = x$ is given by (2). Let us compare SAA and the point-prediction-driven decision (using a random forest) to this optimal decision in the two decision problems presented. Let us also consider our proposals (5)-(8) and others that will be introduced in Section 2.

We consider a particular instance of the two-stage shipment planning problem with $d_z = 5$ warehouses and $d_y = 12$ locations, where we observe some features predictive of demand. In both cases we consider $d_x = 3$ and data $S_N$ that, instead of iid, is sampled from a multidimensional evolving process in order to simulate real-world data collection. We give the particular parameters of the problems in the supplementary Section 13. In Figure 2a, we report the average performance of the various solutions with respect to the *true* distributions.

The full-information optimum clearly does the best with respect to the true distributions, as expected. The SAA and point-prediction-driven decisions have performances that quickly converge to suboptimal values. The former because it does not use observations on $X$ and the latter because it does not take into account the remaining uncertainty after observing $X = x$.[1] In comparison, we find that our proposals converge upon the full-information optimum given sufficient data. In Section 4.3, we study the general asymptotics of our proposals and prove that the convergence observed here empirically is generally guaranteed under only mild conditions.

Inspecting the figure further, it seems that ignoring $X$ and using only the data on $Y$, as SAA does, is appropriate when there is very little data; in both examples, SAA outperforms other data-driven approaches for $N$ smaller than $\sim 64$. Past that point, our constructions of predictive prescriptions, in particular (5)-(8), leverage the auxiliary data effectively and achieve better, and eventually optimal, performance. The predictive prescription motivated by RF is notable in particular for performing no worse than SAA in the small $N$ regime, and better in the large $N$ regime.

In this example, the dimension $d_x$ of the observations $x$ was relatively small at $d_x = 3$. In many practical problems, this dimension may well be bigger, potentially inhibiting performance. E.g., in our real-world application in Section 6, we have $d_x = 91$. To study the effect of the dimension of $x$ on the performance of our proposals, we consider polluting $x$ with additional dimensions of uninformative components distributed as independent normals. The results, shown in Figure 2b, show that while some of the predictive prescriptions show deteriorating performance with growing dimension $d_x$, the predictive prescriptions based on CART and RF are largely unaffected, seemingly able to detect the 3-dimensional subset of features that truly matter. In the supplemental Section 13.2 we also consider an alternative setting of this experiment where additional dimensions carry marginal predictive power.

## 1.2. Relevant Literature

Stochastic optimization as in (1) has long been the focus of decision making under uncertainty in OR/MS problems (cf. Birge and Louveaux (2011)) as has its multi-period generalization known commonly as dynamic programming (cf. Bertsekas (1995)). The solution of stochastic optimization problems as in (1) in the presence of data $\{y^1, \ldots, y^N\}$ on the quantity of interest is a topic of active research. The traditional approach is the sample average approximation (SAA) where the true distribution is replaced by the empirical one (cf. Shapiro (2003), Shapiro and Nemirovski (2005), Kleywegt et al. (2002)). Other approaches include stochastic approximation (cf. Robbins and Monro (1951), Nemirovski et al. (2009)), robust SAA (cf. Bertsimas et al. (2014)), and data-driven mean-variance distributionally-robust optimization (cf. Delage and Ye (2010), Calafiore and El Ghaoui (2006)). A notable alternative approach to decision making under uncertainty in OR/MS problems is robust optimization (cf. Ben-Tal et al. (2009), Bertsimas et al. (2011)) and its data-driven variants (cf. Bertsimas et al. (2013), Calafiore and Campi (2005)). There is also a vast literature on the tradeoff between the collection of data and optimization as informed by data collected so far (cf. Robbins (1952), Lai and Robbins (1985), Besbes and Zeevi (2009)). In all of these methods for data-driven decision making under uncertainty, the focus is on data in the assumed form of iid observations of the parameter of interest $Y$. On the other hand, ML has attached great importance to the problem of supervised learning wherein the conditional expectation (regression) or mode (classification) of target quantities $Y$ given auxiliary observations $X = x$ is of interest (cf. Trevor et al. (2001), Mohri et al. (2012)).

---

[1] Note that the uncertainty of the point prediction in estimating the conditional expectation, gleaned e.g. via the bootstrap, is the wrong uncertainty to take into account, in particular because it shrinks to zero as $N \to \infty$.

Statistical decision theory is generally concerned with the optimal selection of statistical estimators (cf. Berger (1985), Lehmann and Casella (1998)). Following the early work of Wald (1949), a loss function such as sum of squared errors or of absolute deviations is specified and the corresponding admissibility, minimax-optimality, or Bayes-optimality are of main interest. Statistical decision theory and ML intersect most profoundly in the realm of regression via empirical risk minimization (ERM), where a regression model is selected on the criterion of minimizing empirical average of loss. A range of ML methods arise from ERM applied to certain function classes and extensive theory on function-class complexity has been developed to analyze these (cf. Bartlett and Mendelson (2003), Vapnik (2000, 1992)). Such ML methods include ordinary linear regression, ridge regression, the LASSO of Tibshirani (1996), quantile regression, and $\ell_1$-regularized quantile regression of Belloni and Chernozhukov (2011). ERM is also closely connected with $M$-estimation Geer (2000), which estimates a distributional parameter that maximizes an average of a function of the parameter by the estimate that maximizes the corresponding empirical average. Unlike $M$-estimation theory, which is concerned with estimation and inference, ERM theory is only concerned with out-of-sample performance and can be applied more flexibly with less assumptions.

In certain OR/MS decision problems, one can employ ERM to select a decision policy, conceiving of the loss as costs. Indeed, the loss function used in quantile regression is exactly equal to the cost function of the newsvendor problem of inventory management. Rudin and Vahn (2014) consider this loss function and the selection of a univariate-valued linear function with coefficients restricted in $\ell_1$-norm in order to solve a newsvendor problem with auxiliary data, resulting in a method similar to Belloni and Chernozhukov (2011). Kao et al. (2009) study finding a convex combination of two ERM solutions, the least-cost decision and the least-squares predictor, which they find to be useful when costs are quadratic. In more general OR/MS problems where decisions are constrained, we show in the supplemental Section 8 that ERM is not applicable. Even when it is, a linear decision rule may be inappropriate as we show by example. For the limited problems where ERM is applicable, we generalize the standard function-class complexity theory and out-of-sample guarantees to multivariate decision rules since most OR/MS problems involve multivariate decisions.

Instead of ERM, we are motivated more by a strain of non-parametric ML methods based on local learning, where predictions are made based on the mean or mode of past observations that are in some way similar to the one at hand. The most basic such method is $k$NN (cf. Altman (1992)), which define the prediction as a locally constant function depending on which $k$ data points lie closest. A related method is Nadaraya-Watson kernel regression (KR) (cf. Nadaraya (1964), Watson (1964)), which is notable for being highly amenable to theoretical analysis but sees less use in practice. KR weighting for solving conditional stochastic optimization problems as in (2) has been considered in Hanasusanto and Kuhn (2013), Hannah et al. (2010) but these have not considered the more general connection to a great variety of ML methods used in practice and neither have they considered asymptotic optimality rigorously. A more widely used local learning regression method than KR is local regression (Cameron and Trivedi (2005) pg. 311) and in particular the LOESS method of Cleveland and Devlin (1988). Even more widely used are recursive partitioning methods, most often in the form of trees and most notably CART of Breiman et al. (1984). Ensembles of trees, most notably RF of Breiman (2001), are known to be very flexible and have competitive performance in a great range of prediction problems. The former averages locally over a partition designed based on the data (the leaves of a tree) and the latter combines many such averages. While there are many tree-based methods and ensemble methods, we focus on CART and RF because of their popularity and effectiveness in practice.

## 2. From Data to Predictive Prescriptions

Recall that we are interested in the conditional-stochastic optimization problem (2) of minimizing uncertain costs $c(z; Y)$ after observing $X = x$. The key difficulty is that the true joint distribution $\mu_{X,Y}$, which specifies problem (2), is unknown and only data $S_N$ is available. One approach may be to approximate $\mu_{X,Y}$ by the empirical distribution $\hat{\mu}_N$ over the data $S_N$ where each datapoint $(x^i, y^i)$ is assigned mass $1/N$. This, however, will in general fail unless $X$ has small and finite support; otherwise, either $X = x$ has not been observed and the conditional expectation is undefined with respect to $\hat{\mu}_N$ or it has been observed, $X = x = x^i$ for some $i$, and the conditional distribution is a degenerate distribution with a single atom at $y^i$ without any uncertainty. Therefore, we require some way to generalize the data to reasonably estimate the conditional expected costs for any $x$. In some ways this is similar to, but more intricate than, the prediction problem where $\mathbb{E}[Y|X=x]$ is estimated from data for any possible $x \in \mathcal{X}$. We are therefore motivated to consider predictive methods and their adaptation to our cause.

In the next subsections we propose a selection of constructions of predictive prescriptions $\hat{z}_N(x)$, each motivated by a local-learning predictive methodology. All the constructions in this section will take the common form of defining some data-driven weights $w_{N,i}(x)$ and optimizing the decision $\hat{z}_N$ against a re-weighting of the data, as in (3):

$$\hat{z}_N^{\text{local}}(x) \in \arg\min_{z \in \mathcal{Z}} \sum_{i=1}^{N} w_{N,i}(x) c(z; y^i). \tag{11}$$

In some cases the weights are nonnegative and can be understood to correspond to an estimated conditional distribution of $Y$ given $X=x$. But, in other cases, some of the weights may be negative and this interpretation breaks down.

### 2.1. $k$NN

Motivated by $k$-nearest-neighbor regression we propose

$$w_{N,i}^{k\text{NN}}(x) = \tfrac{1}{k}\mathbb{I}\left[x^i \text{ is a } k\text{NN of } x\right], \tag{12}$$

giving rise to the predictive prescription (5). Ties among equidistant data points are broken either randomly or by a lower-index-first rule. Finding the $k$NNs of $x$ without pre-computation can clearly be done in $O(Nd)$ time. Data-structures that speed up the process at query time at the cost of pre-computation have been developed (cf. Bentley (1975)) and there are also approximate schemes that can significantly speed up queries (c.f. Arya et al. (1998)).

### 2.2. Kernel Methods

The Nadaraya-Watson kernel regression (KR; cf. Nadaraya (1964), Watson (1964)) estimates $m(x) = \mathbb{E}[Y|X=x]$ by

$$\hat{m}_N(x) = \frac{\sum_{i=1}^{N} y^i K\left((x^i-x)/h_N\right)}{\sum_{i=1}^{N} K\left((x^i-x)/h_N\right)},$$

where $K : \mathbb{R}^d \to \mathbb{R}$, known as the kernel, satisfies $\int K < \infty$ (and often unitary invariance) and $h_N > 0$, known as the bandwidth. We restrict our attention to the following common kernels: $K(x) = \mathbb{I}\left[\|\|x\| \leq 1\right]$ (Naïve), $K(x) = (1 - \|x\|^2)\mathbb{I}\left[\|\|x\| \leq 1\right]$ (Epanechnikov), and $K(x) = (1 - \|x\|^3)^3\mathbb{I}\left[\|\|x\| \leq 1\right]$ (Tri-cubic). For these (nonnegative) kernels, KR is the result of the conditional distribution estimate that arises from the Parzen-window density estimates (cf. Parzen (1962)) of $\mu_{X,Y}$ and $\mu_X$ (i.e., their ratio). In particular, using the same conditional distribution estimate, the following weights lead to a predictive prescription as in (3):

$$w_{N,i}^{\text{KR}}(x) = \frac{K\left((x^i-x)/h_N\right)}{\sum_{j=1}^{N} K\left((x^j-x)/h_N\right)}. \tag{13}$$

Note that the naïve kernel with bandwidth $h_N$ corresponds directly to uniformly weighting all neighbors of $x$ that are within a radius $h_N$.

A recursive modification to (13) that is motivated by an alternative kernel regressor introduced by Devroye and Wagner (1980) is

$$w_{N,i}^{\text{recursive-KR}}(x) = \frac{K\big((x^i - x)/h_i\big)}{\sum_{j=1}^{N} K\big((x^j - x)/h_j\big)}, \tag{14}$$

where now the bandwidths $h_i$ are selected per-data-point and independent of $N$. From a theoretical point of view, much weaker conditions are necessary to ensure good asymptotic behavior of (14) compared to (13), as we will see in the next section.

### 2.3. Local Linear Methods

Whereas KR estimates $m(x)$ by the best local constant prediction weighted by the kernel (i.e., the weighted average), local linear regression estimates $m(x)$ by the best local linear prediction weighted by the kernel:

$$\hat{m}_N(x) = \arg\min_{\beta_0} \ \min_{\beta_1} \sum_{i=1}^{N} k_i(x) \left(y^i - \beta_0 - \beta_1^T(x^i - x)\right)^2.$$

In prediction, local linear methods are known to be preferable over KR (cf. Fan (1993)). Using this to locally approximate the conditional costs $\mathbb{E}\big[c(z; Y)\big| X = x\big]$ by a linear function we will arrive at a functional estimate and a predictive prescription as in (3) with the weights

$$w_{N,i}^{\text{LOESS}}(x) = \frac{\tilde{w}_{N,i}(x)}{\sum_{j=1}^{N} \tilde{w}_{N,j}(x)}, \quad \tilde{w}_{N,i}(x) = k_i(x)\left(1 - \sum_{j=1}^{n} k_j(x)(x^j - x)^T \Xi(x)^{-1}(x^i - x)\right), \tag{15}$$

where $\Xi(x) = \sum_{i=1}^{n} k_i(x)(x^i - x)(x^i - x)^T$ and $k_i(x) = K\left((x^i - x)/h_N(x)\right)$. In LOESS regression per (Cleveland and Devlin 1988), $K$ is the tri-cubic kernel and $h_N(x)$ is the distance to $x$'s $k$-nearest neighbor with $k$ fixed. In Section 4.1, we establish the computational tractability of predictive prescriptions as in (3) when weights are nonnegative. The weights (15), however, may sometimes be negative. Nonetheless, as $N$ increases, these weights will always become nonnegative. As such, we propose a modification of weights (15) that ensures all weights are nonnegative without sacrificing asymptotic optimality (see Section 4.3):

$$w_{N,i}^{\text{LOESS*}}(x) = \frac{\tilde{w}_{N,i}(x)}{\sum_{j=1}^{N} \tilde{w}_{N,j}(x)}, \quad \tilde{w}_{N,i}(x) = k_i(x) \max\{1 - \sum_{j=1}^{n} k_j(x)(x^j - x)^T \Xi(x)^{-1}(x^i - x), 0\}. \tag{16}$$

### 2.4. Trees

In prediction, CART (Breiman et al. 1984) recursively splits the sample $S_N$ into regions in $\mathcal{X}$ along axis-aligned cuts (one-hot hyperplanes) so to gain reduction in an impurity measure in the response variable $Y$ within each region. Common impurity measures are Gini or entropy for classification and squared error for univariate regression. Multivariate impurity measures are the component-wise average of univariate impurities. Once a tree is constructed, the value of $m(x)$ is estimated by the average of $y^i$'s associated with the $x^i$'s that reside in the same region as $x$.

Regardless of the particular method chosen, a final partition can be represented as a binning rule identifying points in $\mathcal{X}$ with the disjoint regions, $\mathcal{R} : \mathcal{X} \to \{1, \dots, r\}$. The partition is then the disjoint union $\mathcal{R}^{-1}(1) \sqcup \cdots \sqcup \mathcal{R}^{-1}(r) = \mathcal{X}$. The tree regression estimates correspond directly to taking averages over the uniform distribution of the data points residing in the region $R(x)$. For our prescription problem, we propose to use the binning rule to construct weights as follows for a predictive prescription of the form (3):

$$w_{N,i}^{\text{CART}}(x) = \frac{\mathbb{I}\left[\mathcal{R}(x) = \mathcal{R}(x^i)\right]}{\left|\left\{j : R(x^j) = R(x)\right\}\right|}. \tag{17}$$

Notice that the weights (17) are piecewise constant over the partitions and therefore the recommended optimal decision $\hat{z}_N(x)$ is also piecewise constant. Therefore, solving $r$ optimization problems after the recursive partitioning process, the resulting predictive prescription can be fully compiled into a decision tree, with the decisions that are truly decisions. This also retains CART's lauded interpretability.[2]

### 2.5. Ensembles

A random forest (Breiman 2001) is an ensemble of trees each trained a random subsample of the data with random subset of components of $X$ considered at each tree node. After training such a random forest of trees, we can extract the partition rules $\mathcal{R}_t$ $t = 1, \dots, T$, one for each tree in the forest. We propose to use these to construct the following weights as follows for a predictive prescription of the form (3):

$$w_{N,i}^{\text{RF}}(x) = \frac{1}{T} \sum_{t=1}^{T} \frac{\mathbb{I}\left[\mathcal{R}^t(x) = \mathcal{R}^t(x^i)\right]}{\left|\left\{j : R^t(x^j) = R^t(x)\right\}\right|}. \tag{18}$$

In Section 1.1 we demonstrated that our predictive prescription based on RF, given in eq. (8), performed well overall in two different problems, for a range of sample sizes, and for a range of dimensions $d_x$. Based on this evidence of flexible performance, we choose our predictive prescription based on RF for our real-world application, which we study in Section 6.

## 3. From Data to Predictive Prescriptions When Decisions Affect Uncertainty

Up to now, we have assumed that the effect of the decision $z$ on costs is wholly encapsulated in the cost function and that the choice of $z$ does not directly affect the realization of uncertainty $Y$. However, in some settings, such as in the presence of pricing decisions, this assumption clearly does not hold – as one increases a price control, demand diminishes, and the *causal* effect of pricing is not known a priori (e.g., can be abstracted in the cost function) and must be derived from data. In such cases, we must take into account the effect of our decision $z$ on the uncertainty $Y$ by considering historical data $\{(x^1, y^1, z^1), \ldots, (x^N, y^N, z^N)\}$, where we have also recorded historical observations of the variable $Z$, which represents the historical decision taken in each instance. Using potential outcomes, we let $Y(z)$ denote the value of the uncertain variable that would be observed if decision $z$ were chosen. For each data point $i$, only the realization corresponding to the chosen decision $z^i$ is revealed, $y^i = y^i(z^i)$. The counterfactual $y^i(z)$ that would have been observed under any other decision $z \neq z^i$ is not available for measurement. (For detail on potential outcomes and history see Imbens and Rubin 2015, Chapters 1-2.)

---

[2] A more direct application of tree methods to the prescription problem would have us consider the impurities being minimized in each split to be equal to the mean cost $c(z; y)$ of taking the best constant decision $z$ in each side of the split. However, since we must consider splitting on each variable and at each data point to find the best split (cf. pg. 307 of Trevor et al. (2001)), this can be overly computationally burdensome for all but the simplest problems that admit a closed form solution such as least sum of squares or the newsvendor problem.

Since only some parts of our decision may have unknown effects on uncertainty, we decompose our decision variable into the part with unknown effect (e.g., pricing decisions) and known effect (e.g., production decisions) in the following way:

ASSUMPTION 1 (**Decomposition of Decision**). *For some decomposition $z = (z_1, z_2)$ only $z_1 \in \mathbb{R}^{d_{z_1}}$ affects the uncertainty, i.e.,*

$$Y(z_1, z_2) = Y(z_1, z_2') \quad \forall (z_1, z_2), (z_1, z_2') \in \mathcal{Z}.$$

*For brevity, we write $Y(z) = Y(z_1)$. And, we let $\mathcal{Z}_1(z_2) = \{z_1 : (z_1, z_2) \in \mathcal{Z}\}$, $\mathcal{Z}_1 = \{z_1 : \exists z_2 \; (z_1, z_2) \in \mathcal{Z}\}$, $\mathcal{Z}_2(z_1) = \{z_2 : (z_1, z_2) \in \mathcal{Z}\}$, $\mathcal{Z}_2 = \{z_2 : \exists z_1 \; (z_1, z_2) \in \mathcal{Z}\}$.*

For example, in pricing, if $z_1 \in [0, \infty)$ represents a price control for a product and $Y$ represents realized demand, then $\{(z_1, Y(z_1)) : z_1 \in [0, \infty)\}$ represents the *random* demand curve. If in the $i^{\text{th}}$ data point the price was $z_1^i$, then we only observe the single point $(z_1^i, y^i(z_1^i))$ on this random curve. Decision components $z_2$ could represent, for example, a production and shipment plan, which does not affect demand but does affect final costs.

The immediate generalization of problem (2) to this setting is

$$v^*(x) = \min_{z \in \mathcal{Z}} \mathbb{E}\left[c(z; Y(z)) \big| X = x\right], \qquad z^*(x) \in \mathcal{Z}^*(x) = \arg\min_{z \in \mathcal{Z}} \mathbb{E}\left[c(z; Y(z)) \big| X = x\right]. \quad (19)$$

This problem depends on understanding the joint distribution of $(X, Y(z))$ for each $z \in Z$ and, in this full information setting, chooses $z$ for least expected cost given the observation $X = x$ and the effect $z$ would have on the uncertainty $Y(z)$. Assumption 1 allows problem (19) to encompass the standard conditional stochastic optimization problem (2) by letting $z = z_2$ and $d_{z_1} = 0$. On the other hand, Assumption 1 is non-restrictive in the sense that it can be as general as necessary by letting $z = z_1$, i.e., no decomposing into parts of unknown effect and known no effect. For these reason, we maintain the notation $v^*(x)$, $z^*(x)$, $\mathcal{Z}^*(x)$.

Given only the data $(x^i, y^i, z^i)$ on $(X, Y, Z)$ and without any assumptions, problem (19) is in fact not well-specified because of the missing data on the counterfactuals. In particular, for any fixed joint distribution for $(X, Y, Z)$, there are many possible distributions of $(X, Y(z))$ for each $z \in \mathcal{Z}$ that all agree with the same distribution of $(X, Y, Z)$ via the transformation $Y = Y(Z)$ but can each give rise to different optimal solutions $z^*(x)$ in (19) (see Bertsimas and Kallus 2016). Therefore, problem (19) may not be solved using the data alone.

To eliminate this issue, we must make additional assumptions about the data. Here, we make the assumption that controlling for $X$ is sufficient for isolating the effect of $z$ on $Y$.

ASSUMPTION 2 (**Ignorability**). *For every $z \in \mathcal{Z}$, $Y(z)$ is independent of $Z$ conditioned on $X$.*

In words, Assumption 2 says that, historically, $X$ accounts for all the features associated with the instance $\{Y(z) : z \in \mathcal{Z}\}$ that may have influenced managerial decision making. In the causal inference literature, this assumption is standard for ensuring identifiability of causal effects (Rosenbaum and Rubin 1983).

In stark contrast to many situations in causal inference dealing with latent self-selection, Assumption 2 is particularly defensible in our specific setting. In the setting we consider, $Z$ represents historical managerial decisions and, just like future decisions to be made by the learned predictive prescription, these decisions must have been made based on observable quantities available to the manager. As long as these quantities were also recorded as part of $X$ then Assumption 2 is guaranteed to hold. Alternatively, were decisions $Z$ taken at random for exploration then Assumption 2 holds trivially.

### 3.1. Adapting local-learning methods

We now show how to generalize the predictive prescriptions from Section 2 to solve problem (19) when decisions affect uncertainty based on data on $(X, Y, Z)$. We begin with a rephrasing of problem (19) based on Assumptions 1 and 2. The proof is given in the E-companion.

THEOREM 1. *Under Assumptions 1 and 2, problem* (19) *is equivalent to,*

$$\min_{(z_1, z_2) \in \mathcal{Z}} \mathbb{E}\left[ c(z; Y) \big| X = x,\, Z_1 = z_1 \right]. \tag{20}$$

Note that problem (20) depends only on the distribution of the data $(X, Y, Z)$, does not involve unknown counterfactuals, and has the form of a conditional stochastic optimization problem. Correspondingly, all predictive-prescriptive local-learning methods from Section 2 can be adapted to this problem by simply augmenting the data $x^i$ with $z_1^i$. In particular, we can consider data-driven predictive prescriptions of the form

$$\hat{z}_N(x) \in \arg\min_{z \in \mathcal{Z}} \sum_{i=1}^{N} w_{N,i}(x, z_1) c(z; y^i), \tag{21}$$

where $w_{N,i}(x, z_1)$ are weight functions derived from the data by simply taking the same approach as in Section 2 but treating $z_1$ as part of the $X$ data. In particular, for each method in Section 2, we let $\tilde{x}^i = (x^i, z_1^i)$, construct weights $w_{N,i}(\tilde{x})$ based on data $\tilde{S}_N = \{(\tilde{x}^1, y^1), \ldots, (\tilde{x}^N, y^N)\}$, and plug $w_{N,i}(\tilde{x})$ into (21), and compute $\hat{z}_N(x)$. For example, the $k$NN approach applied to (19) has the form (21) with weights

$$w_{N,i}^{k\mathrm{NN}}(x, z_1) = \frac{1}{k}\mathbb{I}\left[(x^i, z_1^i) \text{ is a } k\mathrm{NN of } (x, z_1)\right].$$

As we discuss in Section 4.2, there is an increased computational burden in solving problem (21) when decisions affect uncertainty, compared to our standard predictive prescriptions from Section 2. As we show in Section 4.4, this approach produces prescriptions that are asymptotically optimal even when our decisions have an unknown effect on uncertainty.

### 3.2. Example: two-stage shipment planning with pricing

Consider a pricing variation on our two-stage shipment planning problem from Section 1.1. We introduce an additional decision variable $z_1 \in [0, \infty)$ for the price at which we sell the product. The uncertain demand at the $d_y$ locations $Y(z_1)$ depends on the price we set. In the first stage, we determine price $z_1$ and amounts $z_2$ at $d_{z_2}$ warehouses. In the second stage, instead of shipping from warehouses to satisfy all demand, we can ship as much as we would like. Our profit is the price times number of units sold minus production and transportation costs. Assuming we behave optimally in the second stage, we can write the problem using the cost function and feasible set

$$c(z; y) = p_1 \sum_{i=1}^{d_{z_2}} z_{2,i} + \min_{(t,s) \in Q(z,y)} \left(p_2 \sum_{i=1}^{d_{z_2}} t_i + \sum_{i=1}^{d_{z_2}} \sum_{j=1}^{d_y} (c_{ij} - z_1) s_{ij}\right),$$

$$\mathcal{Z} = \left\{(z_1, z_2) \in \mathbb{R}^{1+d_{z_2}} : z_1, z_2 \geq 0\right\},$$

where $\mathcal{Q}(z, y) = \{(s, t) \in \mathbb{R}^{(d_z \times d_y) \times d_z} : t \geq 0, \ s \geq 0, \ \sum_{i=1}^{d_z} s_{ij} \leq y_j \, \forall j, \ \sum_{j=1}^{d_y} s_{ij} \leq z_{2,i} + t_i \, \forall i\}$.

We now consider observing not only $X$ and $Y$ but also $Z_1$. We consider the same parameters of the problem as in Section 1.1 with an added unknown effect of price on demand so that higher prices induce lower demands. The particular parameters are given in the supplementary Section 13. In Figure 3, we report the average negative profits (production and shipment costs less revenues) of various solutions with respect to the true distributions. We include the full information optimum (19) as well as all of our local-learning methods applied as described in Section 3.1. Again, we compare to SAA and to the point-prediction-driven decision (using a random forest to fit $\hat{m}_N(x, z_1)$, a predictive model based on both $x$ and $z_1$).

We see that our local-learning methods converge upon the full-information optimum as more data becomes available. On the other hand, SAA, which considers only data $y^i$, will always have out-of-sample profits 0 as it will drive $z_1$ to infinity, where demand goes to zero faster than linear. The point-prediction-driven decision performs comparatively well for small $N$, learning quickly the average effect of pricing, but does not converge to the full-information optimum as we gather more data. Overall, our predictive-prescription using RF that addresses the unknown effect of pricing decisions on uncertain demand performs the best.

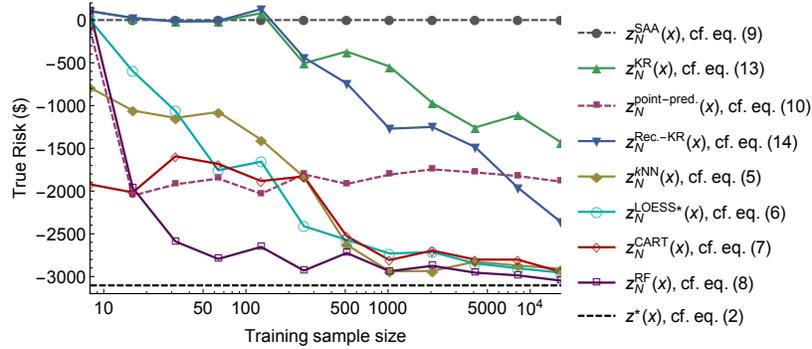

**Figure 3** Performance of various prescriptions in the two-stage shipment planning with pricing problem.

# 4. Properties of Local Predictive Prescriptions

In this section, we study two important properties of local predictive prescriptions: computational tractability and asymptotic optimality. All proofs are given in the E-companion.

## 4.1. Tractability

In Section 2, we considered a variety of predictive prescriptions $\hat{z}_N(x)$ that are computed by solving the optimization problem (3). An important question is then when is this optimization problem computationally tractable to solve. As an optimization problem, problem (3) differs from the problem solved by the standard SAA approach (9) only in the weights given to different observations. Therefore, it is similar in its computational complexity and we can defer to computational studies of SAA such as Shapiro and Nemirovski (2005) to study the complexity of solving problem (3). For completeness, we develop sufficient conditions for problem (3) to be solvable in polynomial time.

THEOREM 2. *Fix $x$ and weights $w_{N,i}(x) \geq 0$. Suppose $\mathcal{Z}$ is a closed convex set and let a separation oracle for it be given. Suppose also that $c(z; y)$ is convex in $z$ for every fixed $y$ and let oracles be given for evaluation and subgradient in $z$. Then for any $x$ we can find an $\epsilon$-optimal solution to (3) in time and oracle calls polynomial in $N_0$, $d$, $\log(1/\epsilon)$ where $N_0 = \sum_{i=1}^{N} \mathbb{I}[w_{N,i}(x) > 0] \leq N$ is the effective sample size.*

Note that all of weights presented in Section 2 are all nonnegative with the exception of local regression (15), which is what led us to their nonnegative modification (16).

## 4.2. Tractability When Decisions Affect Uncertainty

Solving problem (21) with general weights $w_N^i(x, z_1)$ is generally hard as the objective of problem (21) may be non-convex in $z$. In some specific instances we can maintain tractability, while in others we can devise specialized approaches that allow us to solve problem (21) in practice.

In the simplest case, if $\mathcal{Z}_1 = \{z_{11}, \ldots, z_{1b}\}$ is discrete then the problem can simply be solved by optimizing once for each fixed value of $z_1$, letting $z_2$ remain variable.

THEOREM 3. *Fix $x$ and weights $w_{N,i}(x, z_1) \geq 0$. Suppose $\mathcal{Z}_1 = \{z_{11}, \ldots, z_{1b}\}$ is discrete and that $\mathcal{Z}_2(z_{1j})$ is a closed convex set for each $j = 1, \ldots, b$ and let a separation oracle for it be given. Suppose also that $c((z_1, z_2); y)$ is convex in $z_2$ for every fixed $y, z_1$ and let oracles be given for evaluation and subgradient in $z_2$. Then for any $x$ we can find an $\epsilon$-optimal solution to (21) in time and oracle calls polynomial in $N_0, b, d, \log(1/\epsilon)$ where $N_0 = \sum_{i=1}^{N} \mathbb{I}[w_{N,i}(x) > 0] \leq N$ is the effective sample size.*

Note that the convexity in $z_2$ condition is weaker than convexity in $z$, which would be sufficient.

Alternatively, if $\mathcal{Z}_1$ is not discrete, we can approach the problem using discretization, which leads to exponential dependence in $z_1$'s dimension $d_{z_1}$ and the precision $\log(1/\epsilon)$.

THEOREM 4. *Fix $x$ and weights $w_{N,i}(x, z_1) \geq 0$. Suppose $c((z_1, z_2); y)$ is $L$-Lipschitz in $z_1$ for each $z_2 \in \mathcal{Z}_2$, that $\mathcal{Z}_1$ is bounded, and that $\mathcal{Z}_2(z_1)$ is a closed convex set for each $z_1 \in \mathcal{Z}_1$ and let a separation oracle for it be given. Suppose also that $c((z_1, z_2); y)$ is convex in $z_2$ for every fixed $y, z_1$ and let oracles be given for evaluation and subgradient in $z_2$. Then for any $x$ we can find an $\epsilon$-optimal solution to (21) in time and oracle calls polynomial in $N_0, b, d, \log(1/\epsilon), (L/\epsilon)^{d_{z_1}}$ where $N_0 = \sum_{i=1}^{N} \mathbb{I}[w_{N,i}(x) > 0] \leq N$ is the effective sample size.*

Although the exponential dependence in $d_{z_1}$ and super-logarithmic dependence in $1/\epsilon$ appears problematic, this approach works well in practice only for small $d_{z_1}$. For example, we use this approach in our pricing example in Section 3.2, where $d_{z_1} = 1$, to successfully solve many instances of (21).

For the specific case of tree weights, we can discretize the problem *exactly*, leading to a particularly efficient algorithm in practice. Suppose we are given the CART partition rule $\mathcal{R} : \mathcal{X} \times \mathcal{Z}_1 \rightarrow \{1, \ldots, r\}$, then we can solve problem (21) exactly as follows:

1. Let $x$ be given and fix $w_{N,i}^{\mathrm{CART}}(x, z_1) = \frac{\mathbb{I}\left[\mathcal{R}(x, z_1) = \mathcal{R}(x^i, z_1^i)\right]}{\left|\left\{j : R(x^j, z_1^j) = R(x, z_1)\right\}\right|}$.

2. Find the partitions that contain $x$, $\mathcal{J} = \{j : \exists z_1, \ (x, z_1) \in R^{-1}(j)\}$, and compute the constraints on $z_1$ in each part, $\tilde{\mathcal{Z}}_{1j} = \{z_1 : \exists x, \ (x, z_1) \in R^{(-1)}(j)\}$ for $j \in \mathcal{J}$. This is easily done by going down the tree and at each node, if the node queries the value of $x$ we only take the branch that corresponds to the value of our given $x$ and if the node queries the value of a component of $z_1$ then we take both branches and record the constraint on $z_1$ on each side.

3. For each $j \in \mathcal{J}$, solve

$$v_j = \min_{z \in \mathcal{Z}: z \in \tilde{\mathcal{Z}_{1j}}} \sum_{i:R(x^i, z_1^i) = j} c(z; y^i), \qquad z_j = \arg\min_{z \in \mathcal{Z}: z \in \tilde{\mathcal{Z}_{1j}}} \sum_{i:R(x^i, z_1^i) = j} c(z; y^i).$$

   (These can be solved for in advance for each $j = 1, \ldots, r$ to reduce computation at query time.)

4. Let $j(x) = \arg\min_{j \in \mathcal{J}} v_j$ and $\hat{z}_n(x) = z_{j(x)}$.

This procedure solves (21) exactly for weights $w_{N,i}^{\mathrm{CART}}(x, z_1)$.

### 4.3. Asymptotic Optimality

In Section 1.1, we saw that our predictive prescriptions $\hat{z}_N(x)$ converged to the full-information optimum as the sample size $N$ grew. Next, we show that this anecdotal evidence is supported by mathematics and that such convergence is guaranteed under only mild conditions. We define *asymptotic optimality* as the desirable asymptotic behavior for $\hat{z}_N(x)$.

DEFINITION 1. We say that $\hat{z}_N(x)$ is *asymptotically optimal* if, with probability 1, we have that for $\mu_X$-almost-everywhere $x \in \mathcal{X}$,

$$\lim_{N \to \infty} \mathbb{E}\left[c(\hat{z}_N(x); Y) \big| X = x\right] = v^*(x).$$

We say $\hat{z}_N(x)$ is *consistent* if, with probability 1, we have that for $\mu_X$-almost-everywhere $x \in \mathcal{X}$,

$$\lim_{N \to \infty} ||\hat{z}_N(x) - Z^*(x)|| = 0, \text{ where } ||\hat{z}_N(x) - Z^*(x)|| = \inf_{z \in Z^*(x)} ||\hat{z}_N(x) - z||.$$

To a decision maker, asymptotic optimality is the most critical limiting property as it says that decisions implemented will have performance reaching the best possible. Consistency refers to the consistency of $\hat{z}_N(x)$ as a statistical estimator for the full-information optimizer(s) $\mathcal{Z}^*(x)$ and is perhaps less critical for a decision maker but will be shown to hold nonetheless.

Asymptotic optimality and depends on our choice of $\hat{z}_N(x)$, the structure of the decision problem (cost function and feasible set), and on how we accumulate our data $S_N$. The traditional assumption on data collection is that it constitutes an iid process. This is a strong assumption and is often only a modeling approximation. The velocity and variety of modern data collection often means that historical observations do not generally constitute an iid sample in any real-world application. We are therefore motivated to consider an alternative model for data collection, that of mixing processes. These encompass such processes as ARMA, GARCH, and Markov chains, which can correspond to sampling from evolving systems like prices in a market, daily product demands, or the volume of Google searches on a topic. While many of our results extend to such settings via generalized strong laws of large numbers (Walk 2010), we present only the iid case in the main text to avoid cumbersome exposition and defer these extensions to the supplemental Section 9.2. For the rest of the section let us assume that $S_N$ is generated by iid sampling.

As mentioned, asymptotic optimality also depends on the structure of the decision problem. Therefore, we will also require the following conditions.

ASSUMPTION 3 (**Existence**). *The full-information problem* (2) *is well defined:* $\mathbb{E}\left[|c(z; Y)|\right] < \infty$ *for every* $z \in \mathcal{Z}$ *and* $\mathcal{Z}^*(x) \neq \varnothing$ *for almost every* $x$.

ASSUMPTION 4 (**Continuity**). $c(z; y)$ *is equicontinuous in* $z$: *for any* $z \in \mathcal{Z}$ *and* $\epsilon > 0$ *there exists* $\delta > 0$ *such that* $|c(z; y) - c(z'; y)| \leq \epsilon$ *for all* $z'$ *with* $||z - z'|| \leq \delta$ *and* $y \in \mathcal{Y}$.

ASSUMPTION 5 (**Regularity**). $\mathcal{Z}$ *is closed and nonempty and in addition either*

1. $\mathcal{Z}$ is bounded or

2. $\liminf_{||z|| \to \infty} \inf_{y \in \mathcal{Y}} c(z; y) > -\infty$ and for every $x \in \mathcal{X}$, there exists $D_x \subset \mathcal{Y}$ such that $\lim_{||z|| \to \infty} c(z; y) \to \infty$ uniformly over $y \in D_x$ and $\mathbb{P}\left(y \in D_x \,\middle|\, X = x\right) > 0$.

Under these conditions, we have the following sufficient conditions for asymptotic optimality, which are proven as consequences of universal pointwise convergence results of related supervised learning problem of Walk (2010), Hansen (2008).

THEOREM 5 (**kNN**). *Suppose Assumptions 3, 4, and 5 hold. Let $w_{N,i}(x)$ be as in (12) with $k = \min\left\{\lceil CN^\delta \rceil, N - 1\right\}$ for some $C > 0$, $0 < \delta < 1$. Let $\hat{z}_N(x)$ be as in (3). Then $\hat{z}_N(x)$ is asymptotically optimal and consistent.*

THEOREM 6 (**Kernel Methods**). *Suppose Assumptions 3, 4, and 5 hold and that $\mathbb{E}\left[|c(z; Y)| \max\left\{\log|c(z; Y)|, 0\right\}\right] < \infty$ for each $z$. Let $w_{N,i}(x)$ be as in (13) with $K$ being any of the kernels in Section 2.2 and $h_N = CN^{-\delta}$ for $C > 0$, $0 < \delta < 1/d_x$. Let $\hat{z}_N(x)$ be as in (3). Then $\hat{z}_N(x)$ is asymptotically optimal and consistent.*

THEOREM 7 (**Recursive Kernel Methods**). *Suppose Assumptions 3, 4, and 5 hold. Let $w_{N,i}(x)$ be as in (14) with $K$ being the naïve kernel and $h_i = Ci^{-\delta}$ for some $C > 0$, $0 < \delta < 1/(2d_x)$. Let $\hat{z}_N(x)$ be as in (3). Then $\hat{z}_N(x)$ is asymptotically optimal and consistent.*

THEOREM 8 (**Local Linear Methods**). *Suppose Assumptions 3, 4, and 5 hold, that $\mu_X$ is absolutely continuous and has density bounded away from 0 and $\infty$ on the support of $X$ and twice continuously differentiable, and that costs are bounded over $y$ for each $z$ (i.e., $|c(z; y)| \le g(z)$) and twice continuously differentiable. Let $w_{N,i}(x)$ be as in (15) with $K$ being any of the kernels in Section 2.2 and with $h_N = CN^{-\delta}$ for some $C > 0$, $0 < \delta < 1/d_x$. Let $\hat{z}_N(x)$ be as in (3). Then $\hat{z}_N(x)$ is asymptotically optimal and consistent.*

THEOREM 9 (**Nonnegative Local Linear Methods**). *Suppose Assumptions 3, 4, and 5 hold, that $\mu_X$ is absolutely continuous and has density bounded away from 0 and $\infty$ on the support of $X$ and twice continuously differentiable, and that costs are bounded over $y$ for each $z$ (i.e., $|c(z; y)| \le g(z)$) and twice continuously differentiable. Let $w_{N,i}(x)$ be as in (16) with $K$ being any of the kernels in Section 2.2 and with $h_N = CN^{-\delta}$ for some $C > 0$, $0 < \delta < 1/d_x$. Let $\hat{z}_N(x)$ be as in (3). Then $\hat{z}_N(x)$ is asymptotically optimal and consistent.*

Although we do not have firm theoretical results on the asymptotic optimality of the predictive prescriptions based on CART (eq. (7)) and RF (eq. (8)), we have observed them to converge empirically in Section 1.1.

### 4.4. Asymptotic Optimality When Decisions Affect Uncertainty

When decisions affect uncertainty, the condition for asymptotic optimality is subtly different. Under the identity $Y = Y(Z)$, Definiton 1 does not accurately reflect asymptotic optimality and indeed methods that do not account for the unknown effect of the decision (e.g., if we apply our methods without regard to this effect, ignoring data on $Z_1$) will not reach the full-information optimum given by (19). Instead, we would like to ensure that our decisions have optimal cost when taking into account their effect on uncertainty. The desired asymptotic behavior for $\hat{z}_N(x)$ when decisions affect uncertainty is the more general condition given below.

DEFINITION 2. We say that $\hat{z}_N(x)$ is *asymptotically optimal* if, with probability 1, we have that for $\mu_X$-almost-everywhere $x \in \mathcal{X}$, as $N \to \infty$

$$\lim_{N \to \infty} \mathbb{E}\left[c(\hat{z}_N(x); Y(\hat{z}_N(x))) \big| X = x\right] = \min_{z \in \mathcal{Z}} \mathbb{E}\left[c(z; Y(z)) \big| X = x\right].$$

The following theorem establishes asymptotic optimality for our predictive prescription based on either kernel methods, local linear methods, or nonnegative local linear methods as adapted to the case when decisions affect uncertainty. As in Section 3.1, we use $\tilde{x}^i$ to denote $(x^i, z_1^i)$ and $\tilde{S}_N = \{(\tilde{x}^1, y^1), \dots, (\tilde{x}^N, y^N)\}$. To avoid issues of existence, we focus on weak minimizers $\hat{z}_N(x)$ of (21) and on asymptotic optimality.

THEOREM 10. *Suppose Assumptions 1, 2, 3, 4, and 5 (case 1) hold, that $\mu_{(X,Z_1)}$ is absolutely continuous and has density bounded away from 0 and $\infty$ on the support of $X, Z_1$ and twice continuously differentiable, and that costs are bounded over $y$ for each $z$ (i.e., $|c(z; y)| \leq g(z)$) and twice continuously differentiable. Let $w_{N,i}(\tilde{x})$ be as in (13), (15), or (16) applied to $\tilde{S}_N$ with $K$ being any of the kernels in Section 2.2 and with $h_N = CN^{-\delta}$ for some $C > 0$, $0 < \delta < 1/(d_x + d_{z_1})$. Then for any $\epsilon_N \to 0$, any $\hat{z}_N(x)$ that $\epsilon_N$-minimizes (21) (has objective value within $\epsilon_N$ of the infimum) is asymptotically optimal.*

## 5. Metrics of Prescriptiveness

In this section, we develop a relative, unitless measure of the efficacy of a predictive prescription. An absolute measure of efficacy is marginal expected costs,

$$R(\hat{z}_N) = \mathbb{E}\left[\mathbb{E}\left[c(\hat{z}_N(X); Y) \big| X\right]\right] = \mathbb{E}\left[c(\hat{z}_N(X); Y)\right].$$

Given a validation data set $\tilde{S}_{N_v} = ((\tilde{x}^1, \tilde{y}^1), \dots, (\tilde{x}^{N_v}, \tilde{y}^{N_v}))$, we estimate $R(\hat{z}_N)$ as

$$\hat{R}_{N_v}(\hat{z}_N) = \frac{1}{N_v} \sum_{i=1}^{N_v} c(\hat{z}_N(\tilde{x}^i); \tilde{y}^i).$$

If $\tilde{S}_{N_v}$ is disjoint and independent of the training set $S_N$, then this is an out-of-sample estimate that provides an unbiased estimate of $R(\hat{z}_N)$. While an absolute measure allows one to compare two predictive prescriptions for the same problem and data, a relative measure can quantify the overall

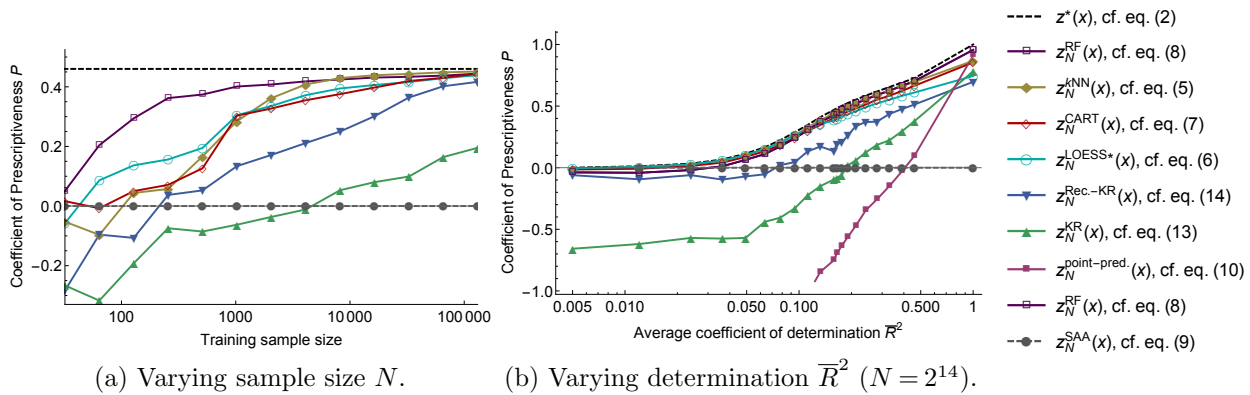

(a) Varying sample size $N$.

(b) Varying determination $\overline{R}^2$ ($N = 2^{14}$).

**Figure 4** **The coefficient of prescriptiveness $P$ in the example from Section 1.1, measured out of sample. The dashed black horizontal line denotes the theoretical limit**

prescriptive content of the data and the efficacy of a prescription on a universal scale. For example, in predictive analytics, the coefficient of determination $R^2$ – rather than the absolute root-mean-squared error – is a unitless quantity used to quantify the overall quality of a prediction and the predictive content of data $X$. $R^2$ measures the fraction of variance of $Y$ reduced, or "explained," by the prediction based on $X$. Another way of interpreting $R^2$ is as the fraction of the way that $X$ and a particular predictive model take us from a data-poor prediction (the sample average) to a perfect-foresight prediction that knows $Y$ in advance.

We define an analogous quantity for the predictive prescription problem, which we term *the coefficient of prescriptiveness*. It involves three quantities. First,

$$\hat{R}_{N_v}(\hat{z}_N) = \frac{1}{N_v} \sum_{i=1}^{N_v} c\left(\hat{z}_N(\tilde{x}^i); \tilde{y}^i\right)$$

is the estimated expected costs due to our predictive prescription. Second,

$$\hat{R}_{N_v}^* = \frac{1}{N_v} \sum_{i=1}^{N_v} \min_{z \in \mathcal{Z}} c\left(z; \tilde{y}^i\right)$$

is the estimated expected costs in the deterministic perfect-foresight counterpart problem, in which one has foreknowledge of $Y$ without any uncertainty (note the difference to the full-information optimum, which does have uncertainty). Third,

$$\hat{R}_{N_v}(z_N^{\text{SAA}}) = \frac{1}{N_v} \sum_{i=1}^{N_v} c\left(\hat{z}_N^{\text{SAA}}; \tilde{y}^i\right) \quad \text{where} \quad \hat{z}_N^{\text{SAA}} \in \arg\min_{z \in \mathcal{Z}} \frac{1}{N} \sum_{i=1}^{N} c\left(z; y^i\right)$$

is the estimated expected costs of a data-driven prescription that is data poor, based only on $Y$ data. This is the SAA solution to the prescription problem, which serves as the analogue to the sample average as a data-poor solution to the prediction problem. Using these three quantities, we define the coefficient of prescriptiveness $P$ as follows:

$$P = 1 - \left(\hat{R}_{N_v}(\hat{z}_N) - \hat{R}_{N_v}^*\right) / \left(\hat{R}_{N_v}(z_N^{\text{SAA}}) - \hat{R}_{N_v}^*\right) \tag{22}$$

The coefficient of prescriptiveness $P$ is a unitless quantity bounded above by 1. A low $P$ denotes that $X$ provides little useful information for the purpose of prescribing an optimal decision in the particular problem at hand or that $\hat{z}_N(x)$ is ineffective in leveraging the information in $X$. A high $P$ denotes that taking $X$ into consideration has a significant impact on reducing costs and that $\hat{z}_N$ is effective in leveraging $X$ for this purpose.

In particular, if $X$ is independent of $Y$ then, under appropriate conditions, $\lim_{N,N_v \to \infty} \hat{R}_{N_v}(z_N^{\mathrm{SAA}}) = \min_{z \in \mathcal{Z}} \mathbb{E}\left[c(z;Y)\right] = \mathbb{E}\left[\min_{z \in \mathcal{Z}} \mathbb{E}\left[c(z;Y)\big|X\right]\right] = \lim_{N,N_v \to \infty} \hat{R}_{N_v}(\hat{z}_N)$, so as $N$ grows, we would see $P$ reach 0. On the other hand, if $Y$ is measurable with respect to $X$, i.e., $Y$ is a function of $X$, then, under appropriate conditions, $\lim_{N,N_v \to \infty} \hat{R}_{N_v}(\hat{z}_N) = \mathbb{E}\left[\min_{z \in \mathcal{Z}} \mathbb{E}\left[c(z;Y)\big|X\right]\right] = \mathbb{E}[\min_{z \in \mathcal{Z}} c(z;Y)] = \lim_{N_v \to \infty} \hat{R}_{N_v}^*$, so as $N$ grows, we would see $P$ reach 1. It is also notable that in the extreme case that $Y$ is function of $X$ then $Y = m(X)$ where $m(x) = \mathbb{E}\left[Y\big|X = x\right]$ so that $\mathbb{E}[\min_{z \in \mathcal{Z}} c(z;Y)] = \mathbb{E}[\min_{z \in \mathcal{Z}} c(z; m(X))]$, and so in this extreme case we would see $P$ reach 1 for $\hat{z}_N^{\mathrm{point\text{-}pred}}$ under appropriate conditions. In the independent case, we would always see $P$ reach a *nonpositive* number under $\hat{z}_N^{\mathrm{point\text{-}pred}}$.

Let us consider the coefficient of prescriptiveness in the example from Section 1.1. For each of our predictive prescriptions and for each $N$, we measure the out of sample $P$ on a validation set of size $N_v = 200$ and plot the results in Figure 4a. Notice that even when we converge to the full-information optimum, $P$ does not approach 1 as $N$ grows. Instead we see that for the same methods that converged to the full-information optimum, we have a $P$ that approaches 0.46. This number represents the extent of the potential that $X$ has to reduce costs in this particular problem. It is the fraction of the way that knowledge of $X$, leveraged correctly, takes us from making a decision under full uncertainty about the value of $Y$ to making a decision in a completely deterministic setting. As is the case with $R^2$, what magnitude of $P$ denotes a successful application depends on the context. In our real-world application in Section 6, we find an out-of-sample $P$ of 0.88.

To consider the relationship between how predictive $X$ is of $Y$ and the coefficient of prescriptiveness, we consider modifying the example by varying the magnitude of residual noise, fixing $N = 2^{14}$. The details are given in the supplementary Section 13. As we vary the noise, we can vary the average coefficient of determination,

$$\overline{R}^2 = 1 - \frac{1}{d_y} \sum_{i=1}^{d_y} \frac{\mathbb{E}[\mathrm{Var}(Y_i|X)]}{\mathrm{Var}(Y_i)},$$

from 0 to 1. In the original example, $\overline{R}^2 = 0.16$. We plot the results in Figure 4b, noting that the behavior matches our description of the extremes above. In particular, when $X$ and $Y$ are independent ($\overline{R}^2 = 0$), we see most methods having a zero coefficient of prescriptiveness, less successful methods (KR) have a somewhat negative coefficient, and the point-prediction-driven decision has a very negative coefficient. When $Y$ is measurable with respect to $X$ ($\overline{R}^2 = 1$), the coefficient of the optimal decision reaches 1, most methods have a coefficient near 1, and the point-prediction-driven decision also has a coefficient near 1 and beats most other methods. While neither extreme is reasonable in practice, throughout the range, the predictive prescription motivated by RF performs particularly well.

# 6.    A Real-World Application

In this section, we apply our approach to a real-world problem faced by the distribution arm of an international media conglomerate (the vendor) and demonstrate that our approach, combined with extensive data collection, leads to significant advantages. The vendor has asked us to keep its identity confidential as well as data on sale figures and specific retail locations. Some figures are therefore shown on relative scales.

## 6.1.    Problem Statement

The vendor sells over 0.5 million entertainment media titles on CD, DVD, and BluRay at over 50,000 retailers across the US and Europe. On average they ship 1 billion units in a year. The retailers range from electronic home goods stores to supermarkets, gas stations, and convenience stores. These have vendor-managed inventory (VMI) and scan-based trading (SBT) agreements with the vendor. VMI means that the inventory is managed by the vendor, including replenishment (which they perform weekly) and planogramming. SBT means that the vendor owns all inventory until sold to the consumer, at which point the retailer buys the unit from the vendor and sells it to the consumer. This means that retailers have no cost of capital in holding the vendor's inventory.

The cost of a unit of entertainment media consists mainly of the cost of production of the content. Media-manufacturing and delivery costs are secondary in effect. Therefore, the primary objective of the vendor is simply to sell as many units as possible and the main limiting factor is inventory capacity at the retail locations. For example, at many of these locations, shelf space for the vendor's entertainment media is limited to an aisle endcap display and no back-of-the-store storage is available. Thus, the main loss incurred in over-stocking a particular product lies in the loss of potential sales of another product that sold out but could have sold more. In studying this problem, we will restrict our attention to the replenishment and sale of video media only and to retailers in Europe.

Apart from the limited shelf space the other main reason for the difficulty of the problem is the particularly high uncertainty inherent in the initial demand for new releases. Whereas items that have been sold for at least one period have a somewhat predictable decay in demand, determining where demand for a new release will start is a much less trivial task. At the same time, new releases present the greatest opportunity for high demand and many sales.

We now formulate the full-information problem. Let $r = 1, \ldots, R$ index the locations, $t = 1, \ldots, T$ index the replenishment periods, and $j = 1, \ldots, d$ index the products. Denote by $z_j$ the order quantity decision for product $j$, by $Y_j$ the uncertain demand for product $j$, and by $K_r$ the overall inventory capacity at location $r$. Considering only the *main* effects on revenues and costs as discussed in the previous paragraph, the problem decomposes on a per-replenishment-period, per-location basis. We therefore wish to solve, for each $t$ and $r$, the following problem:

$$v^*(x_{tr}) = \max \quad \mathbb{E}\left[\sum_{j=1}^d \min\{Y_j, z_j\} \,\middle|\, X = x_{tr}\right] = \sum_{j=1}^d \mathbb{E}\left[\min\{Y_j, z_j\} \,\middle|\, X_j = x_{tr}\right] \qquad (23)$$
$$\text{s.t.} \quad z \geq 0, \ \sum_{j=1}^d z_j \leq K_r,$$

where $x_{tr}$ denotes auxiliary data available at the beginning of period $t$ in the $(t,r)^{\text{th}}$ problem.

Note that had there been no capacity constraint in problem (23) and a per-unit ordering cost were added, the problem would decompose into $d$ separate newsvendor problems, the solution to each being exactly a quantile regression on the regressors $x_{tr}$. As it is, the problem is coupled, but, fixing $x_{tr}$, the capacity constraint can be replaced with an equivalent per-unit ordering cost $\lambda$ via Lagrangian duality and the optimal solution is attained by setting each $z_j$ to the $\lambda^{\text{th}}$ conditional quantile of $Y_j$. However, the reduction to quantile regression does not hold since the dual optimal value of $\lambda$ depends *simultaneously* on all of the conditional distributions of $Y_j$ for $j = 1, \dots, d$.

## 6.2. Applying Predictive Prescriptions to Censored Data

In applying our approach to problem (23), we face the issue that we have data on sales, not demand. That is, our data on the quantity of interest $Y$ is right-censored. In this section, we develop a modification of our approach to correct for this. The results in this section apply generally.

Suppose that instead of data $\{y^1, \dots, y^N\}$ on $Y$, we have data $\{u^1, \dots, u^N\}$ on $U = \min\{Y, V\}$ where $V$ is an observable random threshold, data on which we summarize via $\delta = \mathbb{I}[U < V]$. For example, in our application, $V$ is the on-hand inventory level at the beginning of the period. Overall, our data consists of $\tilde{S}_N = \{(x^1, u^1, \delta^1), \dots, (x^N, u^N, \delta^N)\}$.

One way to deal with this is by considering decisions (sock levels) as affecting uncertainty (sales). As long as demand and threshold are conditionally independent given $X$, Assumption 2 will be satisfied and we can use the approach (21) developed in Section 3.1. However, the particular setting of censored data has a lot structure where we actually know the mechanism of *how* decision affect uncertainty. This allows us to develop a special-purpose solution that side-steps the need to learn the structure of this dependence and computationally less tractable approaches (Section 4.2).

In order to correct for the fact that our observations are in fact censored, we develop a conditional variant of the Kaplan-Meier method (cf. Kaplan and Meier (1958), Huh et al. (2011)) to transform our weights appropriately. Let $(i)$ denote the ordering $u^{(1)} \leq \cdots \leq u^{(N)}$. Given the weights $w_{N,i}(x)$ generated based on the naïve assumption that $y^i = u^i$, we transform these into the weights

$$w_{N,(i)}^{\text{Kaplan-Meier}}(x) = \mathbb{I}\left[\delta^{(i)} = 1\right] \left(\frac{w_{N,(i)}(x)}{\sum_{\ell=i}^N w_{N,(\ell)}(x)}\right) \prod_{k \leq i-1 \,:\, \delta^{(k)}=1} \left(\frac{\sum_{\ell=k+1}^N w_{N,(\ell)}(x)}{\sum_{\ell=k}^N w_{N,(\ell)}(x)}\right). \qquad (24)$$

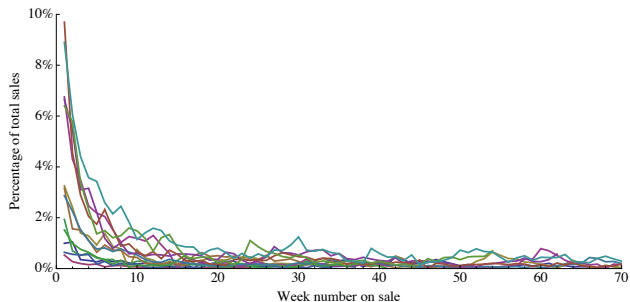

**Figure 5** The percentage of all sales in the German state of Berlin taken up by each of 13 selected titles, starting from the point of release of each title to HE sales.

We next show that the transformation (24) preserves asymptotic optimality under certain conditions. The proof is in the E-companion.

THEOREM 11. *Suppose that $Y$ and $V$ are conditionally independent given $X$, that $Y$ and $V$ share no atoms, that for every $x \in \mathcal{X}$ the upper support of $V$ given $X = x$ is greater than the upper support of $Y$ given $X = x$, and that costs are bounded over $y$ for each $z$ (i.e., $|c(z; y)| \leq g(z)$). Let $w_{N,i}(x)$ be as in (12), (13), (14), (15), or (16) and suppose the corresponding assumptions of Theorem 5, 6, 7, (15), or (16) apply. Let $\hat{z}_N(x)$ be as in (3) but using the transformed weights (24). Then $\hat{z}_N(x)$ is asymptotically optimal and consistent.*

The assumption that $Y$ and $V$ share no atoms (which holds in particular if either is continuous) provides that $\delta \overset{a.s.}{=} \mathbb{I}[Y \leq V]$ so that the event of censorship is observable. In applying this to problem (23), the assumption that $Y$ and $V$ are conditionally independent given $X$ will hold if $X$ captures at least all of the information that past stocking decisions, which are made before $Y$ is realized, may have been based on. The assumption on bounded costs applies to problem (23) because the cost (negative of the objective) is bounded in $[-K_r, 0]$.

## 6.3. Data

In this section, we describe the data collected. To get at the best data-driven predictive prescription, we combine both internal company data and public data harvested from online sources. The predictive power of such public data has been extensively documented in the literature (cf. Asur and Huberman (2010), Choi and Varian (2012), Goel et al. (2010), Da et al. (2011), Gruhl et al. (2005, 2004), Kallus (2014)). Here we study its prescriptive power.

**Internal Data.** The internal company data consists of 4 years of sale and inventory records across the network of retailers, information about each of the locations, and information about each of the items.

We aggregate the sales data by week (the replenishment period of interest) for each feasible combination of location and item. As discussed above, these sales-per-week data constitute a right-censored observation of weekly demand, where censorship occurs when an item is sold out. We

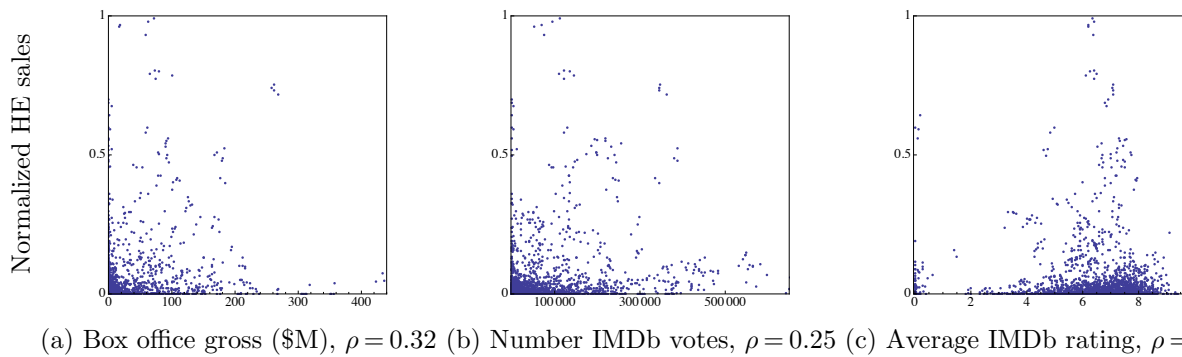

(a) Box office gross ($M), $\rho = 0.32$ (b) Number IMDb votes, $\rho = 0.25$ (c) Average IMDb rating, $\rho = 0.02$

**Figure 6    Scatter plots of various data from IMDb and RT (horizontal axes) against total European sales during first week of HE release (vertical axes, rescaled to anonymize) and corresponding coefficients of correlation ($\rho$).**

developed the transformed weights (24) to tackle this issue exactly. Figure 5 shows the sales life cycle of a selection of titles in terms of their marketshare when they are released to home entertainment (HE) sales and onwards. Since new releases can attract up to almost 10% of sales in their first week of release, they pose a great sales opportunity, but at the same time significant demand uncertainty. Information about retail locations includes to which chain a location belongs and the address of the location. To parse the address and obtain a precise position of the location, including country and subdivision, we used the Google Geocoding API (Application Programming Interface).[3] Information about items include the medium (e.g. DVD or BluRay) and an item "title." The title is a short descriptor composed by a local marketing team in charge of distribution and sales in a particular region and may often include information beyond the title of the underlying content. For example, a hypothetical film titled *The Film* sold in France may be given the item title "THE FILM DVD + LIVRET - EDITION FR", implying that the product is a French edition of the film, sold on a DVD, and accompanied by a booklet (*livret*), whereas the same film sold in Germany on BluRay may be given the item title "FILM, THE (2012) - BR SINGLE", indicating it is sold on a single BluRay disc.

**Public Data: Item Metadata, Box Office, and Reviews.** We sought to collect additional data to characterize the items and how desirable they may be to consumers. For this we turned to the Internet Movie Database (IMDb; `www.imdb.com`) and Rotten Tomatoes (RT; `www.rottentomatoes.com`). IMDb is an online database of information on films and TV series. RT is a website that aggregates professional reviews from newspapers and online media, along with user ratings, of films and TV series.

In order to harvest information from these sources on the items being sold by the vendor, we first had to disambiguate the item entities and extract original content titles from the item titles.

---

[3] See `https://developers.google.com/maps/documentation/geocoding` for details.

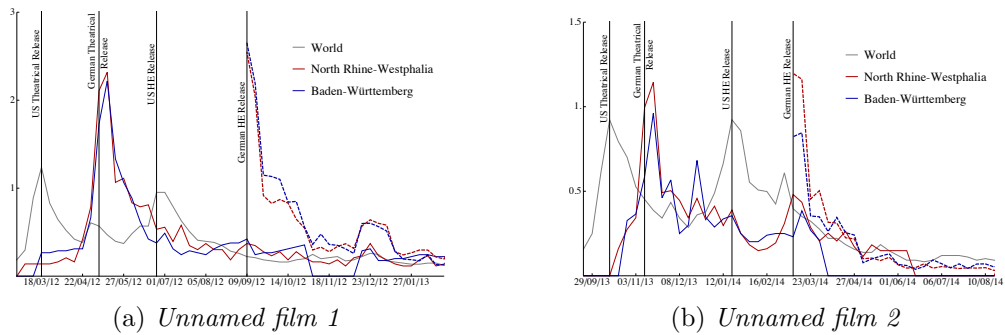

(a) *Unnamed film 1*        (b) *Unnamed film 2*

**Figure 7**    **Weekly search engine attention for two unnamed films in the world and in two populous German states (solid lines) and weekly HE sales for the same films in the same states (dashed lines). Search engine attention and sales are both shown relative to corresponding overall totals in the respective region. The scales are arbitrary but common between regions and the two plots.**

Having done so, we extract the following information from IMDb: type (film, TV, other/unknown); US original release date of content (e.g. in theaters); average IMDb user rating (0-10); number of IMDb users voting on rating; number of awards (e.g. Oscars for films, Emmys for TV) won and number nominated for; the main actors (i.e., first-billed); plot summary (30-50 words); genre(s) (of 26; can be multiple); and MPAA rating (e.g. PG-13, NC-17) if applicable. And the following information from RT: professional reviewers' aggregate score; RT user aggregate rating; number of RT users voting on rating; and if a film, then American box office gross when shown in theaters.

In Figure 6, we provide scatter plots of some of these attributes against sale figures in the first week of HE release. Notice that the number of users voting on the rating of a title is much more indicative of HE sales than the quality of a title as reported in the aggregate score of these votes.

**Public Data: Search Engine Attention.** In the above, we saw that box office gross is reasonably informative about future HE sale figures. The box office gross we are able to access, however, is for the American market and is also missing for various European titles. We therefore would like additional data to quantify the attention being given to different titles and to understand the local nature of such attention. For this we turned to Google Trends (GT; `www.google.com/trends`).[4]

For each title, we measure the relative Google search volume for the search term equal to the original content title in each week from 2011 to 2014 (inclusive) over the whole world, in each European country, and in each country subdivision (states in Germany, cantons in Switzerland, autonomous communities in Spain, etc.). In each such region, after normalizing against the volume of our baseline query, the measurement can be interpreted as the fraction of Google searches for the title in a given week out of all searches in the region, measured on an arbitrary but (approximately) common scale between regions.

---

[4] While GT is available publicly online, access to massive-scale querying and week-level trends data is not public. See acknowledgements.

In Figure 7, we compare this search engine attention to sales figures in Germany for two unnamed films.[5] Comparing panel (a) and (b), we first notice that the overall scale of sales correlates with the overall scale of *local* search engine attention at the time of theatrical release, whereas the global search engine attention is less meaningful (note vertical axis scales, which are common between the two figures). Looking closer at differences between regions in panel (b), we see that, while showing in cinemas, unnamed film 2 garnered more search engine attention in North Rhine-Westphalia (NW) than in Baden-Württemberg (BW) and, correspondingly, HE sales in NW in the first weeks after HE release were greater than in BW. In panel (a), unnamed film 1 garnered similar search engine attention in both NW and BW and similar HE sales as well. In panel (b), we see that the search engine attention to unnamed film 2 in NW accelerated in advance of the HE release, which was particularly successful in NW. In panel (a), we see that a slight bump in search engine attention 3 months into HE sales corresponded to a slight bump in sales. These observations suggest that local search engine attention both at the time of local theatrical release and in recent weeks may be indicative of future sales volumes.

### 6.4. Constructing Auxiliary Data Features and a Random Forest Prediction

For each instance $(t, r)$ of problem (23) and for each item $i$ we construct a vector of numeric predictive features $x_{tri}$ that consist of backward cumulative sums of the sale volume of the item $i$ at location $r$ over the past 3 weeks (as available; e.g., none for new releases), backward cumulative sums of the total sale volume at location $r$ over the past 3 weeks, the overall mean sale volume at location $r$ over the past 1 year, the number of weeks since the original release date of the content (e.g., for a new release this is the length of time between the premier in theaters to release on DVD), an indicator vector for the country of the location $r$, an indicator vector for the identity of chain to which the location $r$ belongs, the total search engine attention to the title $i$ over the first two weeks of local theatrical release globally, in the country, and in the country-subdivision of the location $r$, backward cumulative sums of search engine attention to the title $i$ over the past 3 weeks globally, in the country, and in the country-subdivision of the location $r$, and features capturing item information harvested from IMDb and RT.

Much of the information harvested from IMDb and RT is unstructured in that it is not numeric features, such as plot summaries, MPAA ratings, and actor listings. To capture this information as numerical features that can be used in our framework, we use a range of clustering and community-detection techniques, which we describe in full in supplementary Section 14.

We end up with $d_x = 91$ numeric predictive features. Having summarized these numerically, we train a RF of 500 trees to predict sales. In training the RF, we normalize each the sales in each

---

[5] These films must remain unnamed because a simple search can reveal their European distributor and hence the vendor who prefers their identity be kept confidential.

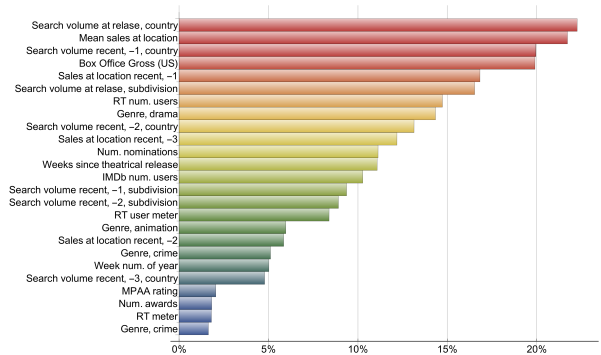

(a) Top 25 $x$ variables in predictive importance, measured as the average over trees of the change in mean-squared error of the tree as percentage of total variance when the value of the variables is randomly permuted among the out-of-bag training data.

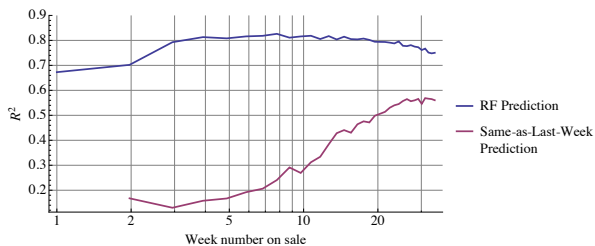

(b) Out-of-sample coefficients of determination $R^2$ for predicting demand next week at different stages of product life cycle.

instance by the training-set average sales in the corresponding location; we de-normalize after predicting. To capture the decay in demand from time of release in stores, we train a separate RFs for sale volume on the $k^{\text{th}}$ week on the shelf for $k = 1, \ldots, 35$ and another RF for the "steady state" weekly sale volume after 35 weeks.

For $k = 1$, we are predicting the demand for a new release, the uncertainty of which, as discussed in Section 6.1, constitutes one of the greatest difficulties of the problem to the company. In terms of predictive quality, when measuring out-of-sample performance we obtain an $R^2 = 0.67$ for predicting sale volume for new releases. The 25 most important features in this prediction are given in Figure 8a. In Figure 8b, we show the $R^2$ obtained also for predictions at later times in the product life cycle, compared to the performance of a baseline heuristic that always predicts for next week the demand of last week.

Considering the uncertainty associated with new releases, we feel that this is a positive result, but at the same time what truly matters is the performance of the prescription in the problem. We discuss this next.

### 6.5. Applying Our Predictive Prescriptions to the Problem

In the last section we discussed how we construct RFs to predict sales, but our problem of interest is to prescribe order quantities. To solve our problem (23), we use the trees in the forests we trained to construct weights $w_{N,i}(x)$ exactly as in (18), then we transform these as in (24), and finally we prescribe data-driven order quantities $\hat{z}_N(x)$ as in (8). Thus, we use our data to go from an observation $X = x$ of our varied auxiliary data directly to a replenishment decision on order quantities.

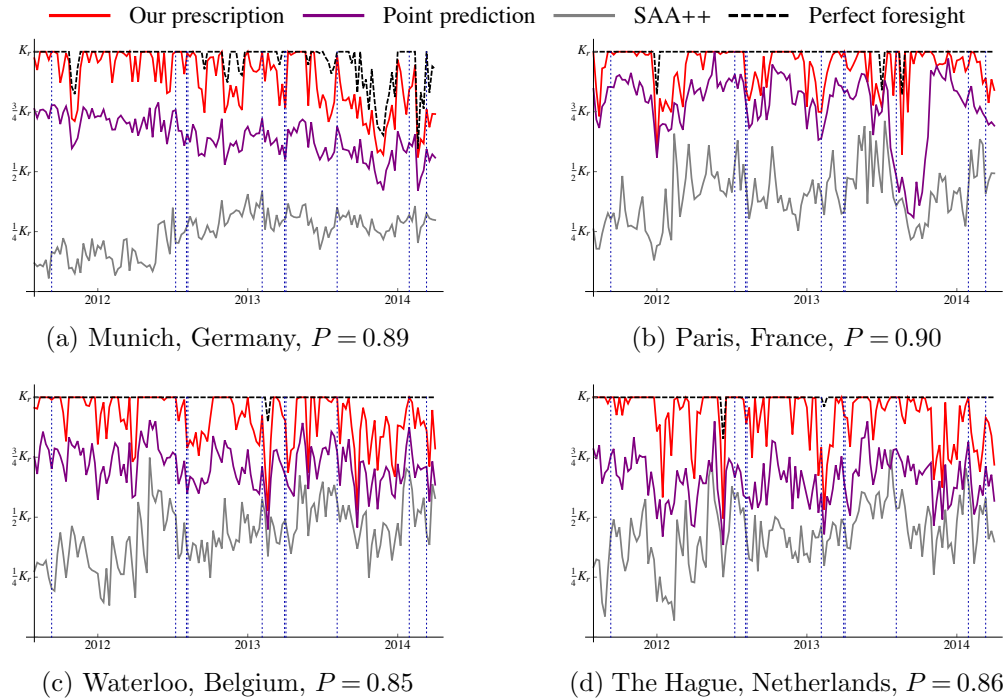

(a) Munich, Germany, $P = 0.89$      (b) Paris, France, $P = 0.90$

(c) Waterloo, Belgium, $P = 0.85$      (d) The Hague, Netherlands, $P = 0.86$

**Figure 9**    The performance of our prescription over time. Blue vertical dashes indicate major release dates. The vertical axis is shown in terms of the location's capacity, $K_r$.

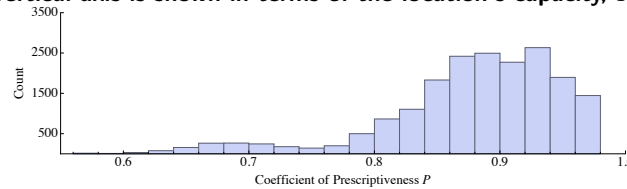

**Figure 10**    The distribution of coefficients of prescriptiveness $P$ over retail locations.

We would like to test how well our prescription does out-of-sample and as an actual live policy. To do this we consider what we would have done over the 150 weeks from December 19, 2011 to November 9, 2014 (inclusive). At each week, we consider only data from time prior to that week, train our RFs on this data, and apply our prescription to the current week. Then we observe what had actually materialized and score our performance.

There is one issue with this approach to scoring: our historical data only consists of sales, not demand. While we corrected for the adverse effect of demand censorship on our prescriptions using the transformation (24), we are still left with censored demand when scoring performance as described above. In order to have a reasonable measure of how good our method is, we therefore consider the problem (23) with capacities $K_r$ that are a *quarter* of their nominal values. In this way, demand censorship hardly ever becomes an issue in the scoring of performance. To be clear, this correction is necessary just for a counterfactual scoring of performance; not in practice.

The transformation (24) already corrects for prescriptions trained on censored observations of the quantity $Y$ that affects true costs.

We compare the performance of our method with three other quantities. One is the performance of the perfect-forecast policy, which knows future demand exactly (no distributions). Another is the performance of a data-driven policy without access to the auxiliary data (i.e., SAA). Because the decay of demand over the lifetime of a product is significant, to make it a fair comparison we let this policy depend on the distributions of product demand based on how long its been on the market. That is, it is based on $T$ separate datasets where each consists of the demands for a product after $t$ weeks on the market (again, considering only past data). Due to this handicap we term it SAA++ henceforth. The last benchmark is the performance of a point-prediction-driven policy using the RF sale prediction. Because there are a multitude of optimal solutions $z_j$ to (23) if we were to let $Y_j$ be deterministic and fixed as our prediction $\hat{m}_{N,j}(x)$, we have to choose a particular one for the point-prediction-driven decision. The one we choose sets order levels to match demand and scale to satisfy the capacity constraint: $\hat{z}_{N,j}^{\text{point-pred}}(x) = K_r \max\{0, \hat{m}_{N,j}(x)\} / \sum_{j'=1}^{d} \max\{0, \hat{m}_{N,j'}(x)\}$.

The ratio of the difference between our performance and that of the prescient policy and the difference between the performance of SAA++ and that of the prescient policy is the coefficient of prescriptiveness $P$. When measured out-of-sample over the 150-week period as these policies make live decisions, we get $P = 0.88$. Said another way, in terms of our objective (sales volumes), our data $X$ and our prescription $\hat{z}_N(x)$ gets us 88% of the way from the best data-poor decision to the impossible perfect-foresight decision. This is averaged over just under 20,000 locations.

In Figure 9, we plot the performance over time at four specific locations, the city of which is noted. Blue vertical dashes in each plot indicate the release dates of the 10 biggest first-week sellers in each location, which turn out to be the same. Two pairs of these coincide on the same week. The plots show a general ordering of performance with our policy beating the point-prediction-driven policy (but not always as seen in a few days in Figure 9b), which in turn beats SAA++ (but not always as seen in a few days in Figure 9d). The $P$ of our policy specific to these locations are 0.89, 0.90, 0.85, and 0.86. The corresponding $P$ of the point-prediction-driven policy are 0.56, 0.57, 0.50, 0.40. That the point-prediction-driven policy outperforms SAA++ (even with the handicap) and provides a significant improvement as measured by $P$ can be attributed to the informativeness of the data collected in Section 6.3 about demand. On most major release dates, the point-prediction-driven policy does relatively worse, which can be attributed to the fact that demand for new releases has the greatest amount of (residual) uncertainty, which the point-prediction-driven policy ignores. When we leverage this data in a manner appropriate for inventory management using our approach, we nearly double the improvement. We also see that on most major release dates, our policy seizes the opportunity to match the perfect foresight performance, but on a few it falls short. In Figure 10, we plot the overall distribution of $P$ of our policy over all retail locations in Europe.

# 7. Concluding Remarks

In this paper, we combine ideas from ML and OR/MS in developing a framework, along with specific methods, for using data to prescribe optimal decisions in OR/MS problems that leverage auxiliary observations. We motivate our methods based on existing predictive methodology from ML, but, in the OR/MS tradition, focus on the making of a decision and on the effect on costs, revenues, and risk. Our approach is generally applicable, tractable, asymptotically optimal, and leads to substantive and measurable improvements in a real-world context.

We feel that the above qualities, together with the growing availability of data and in particular auxiliary data in OR/MS applications, afford our proposed approach a potential for substantial impact in the practice of OR/MS.

## Acknowledgments

The authors would like to thank the anonymous reviewers and associate editor for their very helpful suggestions that helped us improve the paper. We would like to thank Ross Anderson and David Nachum of Google for assistance in obtaining access to Google Trends data. This paper is based upon work supported by the National Science Foundation Graduate Research Fellowship under Grant No. 1122374.

## References

Altman, Naomi S. 1992. An introduction to kernel and nearest-neighbor nonparametric regression. *The American Statistician* **46**(3) 175–185.

Arya, Sunil, David Mount, Nathan Netanyahu, Ruth Silverman, Angela Wu. 1998. An optimal algorithm for approximate nearest neighbor searching in fixed dimensions. *J. ACM* **45**(6) 891–923.

Asur, Sitaram, Bernardo Huberman. 2010. Predicting the future with social media. *WI-IAT*. 492–499.

Bartlett, Peter, Shahar Mendelson. 2003. Rademacher and gaussian complexities: Risk bounds and structural results. *J. Mach. Learn. Res.* **3** 463–482.

Belloni, Alexandre, Victor Chernozhukov. 2011. ℓ1-penalized quantile regression in high-dimensional sparse models. *The Annals of Statistics* **39**(1) 82–130.

Ben-Tal, Aharon, Laurent El Ghaoui, Arkadi Nemirovski. 2009. *Robust optimization*. Princeton University Press.

Bentley, Jon. 1975. Multidimensional binary search trees used for associative searching. *Commun. ACM* **18**(9) 509–517.

Beran, Rudolf. 1981. Nonparametric regression with randomly censored survival data. Tech. rep., Technical Report, Univ. California, Berkeley.

Berger, James O. 1985. *Statistical decision theory and Bayesian analysis*. Springer.

Bertsekas, Dimitri, Angelia Nedić, Asuman Ozdaglar. 2003. *Convex analysis and optimization*. Athena Scientific, Belmont.

Bertsekas, Dimitri P. 1995. *Dynamic programming and optimal control*. Athena Scientific Belmont, MA.

Bertsekas, Dimitri P. 1999. *Nonlinear programming*. Athena Scientific, Belmont.

Bertsimas, Dimitris, David B Brown, Constantine Caramanis. 2011. Theory and applications of robust optimization. *SIAM review* **53**(3) 464–501.

Bertsimas, Dimitris, Vishal Gupta, Nathan Kallus. 2013. Data-driven robust optimization Submitted to *Operations Research*.

Bertsimas, Dimitris, Vishal Gupta, Nathan Kallus. 2014. Robust saa Submitted to *Mathematical Programming*.

Bertsimas, Dimitris, Nathan Kallus. 2016. Pricing from observational data .

Besbes, Omar, Assaf Zeevi. 2009. Dynamic pricing without knowing the demand function: Risk bounds and near-optimal algorithms. *Operations Research* **57**(6) 1407–1420.

Billingsley, P. 1999. *Convergence of Probability Measures*. Wiley, New York.

Birge, John R, Francois Louveaux. 2011. *Introduction to stochastic programming*. Springer.

Blondel, Vincent D, Jean-Loup Guillaume, Renaud Lambiotte, Etienne Lefebvre. 2008. Fast unfolding of communities in large networks. *Journal of Statistical Mechanics: Theory and Experiment* **2008**(10) P10008.

Bradley, Richard. 1986. Basic properties of strong mixing conditions. *Dependence in Probability and Statistics*. Birkhausser, 165–192.

Bradley, Richard. 2005. Basic properties of strong mixing conditions. A survey and some open questions. *Probab. Surv.* **2**(107-44) 37.

Breiman, Leo. 2001. Random forests. *Machine learning* **45**(1) 5–32.

Breiman, Leo, Jerome Friedman, Charles Stone, Richard Olshen. 1984. *Classification and regression trees*. CRC press.

Calafiore, Giuseppe, Marco C Campi. 2005. Uncertain convex programs: randomized solutions and confidence levels. *Mathematical Programming* **102**(1) 25–46.

Calafiore, Giuseppe Carlo, Laurent El Ghaoui. 2006. On distributionally robust chance-constrained linear programs. *Journal of Optimization Theory and Applications* **130**(1) 1–22.

Cameron, A Colin, Pravin K Trivedi. 2005. *Microeconometrics: methods and applications*. Cambridge university press.

Carrasco, Marine, Xiaohong Chen. 2002. Mixing and moment properties of various garch and stochastic volatility models. *Econometric Theory* **18**(1) 17–39.

Choi, Hyunyoung, Hal Varian. 2012. Predicting the present with google trends. *Econ. Rec.* **88**(s1) 2–9.

Cleveland, William S, Susan J Devlin. 1988. Locally weighted regression: an approach to regression analysis by local fitting. *Journal of the American Statistical Association* **83**(403) 596–610.

Da, Zhi, Joseph Engelberg, Pengjie Gao. 2011. In search of attention. *J. Finance* **66**(5) 1461–1499.

Delage, E, Y Ye. 2010. Distributionally robust optimization under moment uncertainty with application to data-driven problems. *Operations Research* **55**(3) 98–112.

Devroye, Luc P, TJ Wagner. 1980. On the l 1 convergence of kernel estimators of regression functions with applications in discrimination. *Zeitschrift für Wahrscheinlichkeitstheorie und verwandte Gebiete* **51**(1) 15–25.

Doukhan, Paul. 1994. *Mixing: Properties and Examples*. Springer.

Dudley, Richard M. 2002. *Real analysis and probability*, vol. 74. Cambridge University Press, Cambridge.

Fan, Jianqing. 1993. Local linear regression smoothers and their minimax efficiencies. *The Annals of Statistics* 196–216.

Geer, Sara A. 2000. *Empirical Processes in M-estimation*, vol. 6. Cambridge university press.

Goel, Sharad, Jake Hofman, Sébastien Lahaie, David Pennock, Duncan Watts. 2010. Predicting consumer behavior with web search. *PNAS* **107**(41) 17486–17490.

Grotschel, M, L Lovasz, A Schrijver. 1993. *Geometric algorithms and combinatorial optimization*. Springer, New York.

Gruhl, Daniel, Laurent Chavet, David Gibson, Jörg Meyer, Pradhan Pattanayak, Andrew Tomkins, J Zien. 2004. How to build a WebFountain: An architecture for very large-scale text analytics. *IBM Syst. J.* **43**(1) 64–77.

Gruhl, Daniel, Ramanathan Guha, Ravi Kumar, Jasmine Novak, Andrew Tomkins. 2005. The predictive power of online chatter. *SIGKDD*. 78–87.

Hanasusanto, Grani Adiwena, Daniel Kuhn. 2013. Robust data-driven dynamic programming. *Advances in Neural Information Processing Systems*. 827–835.

Hannah, Lauren, Warren Powell, David M Blei. 2010. Nonparametric density estimation for stochastic optimization with an observable state variable. *Advances in Neural Information Processing Systems*. 820–828.

Hansen, Bruce E. 2008. Uniform convergence rates for kernel estimation with dependent data. *Econometric Theory* **24**(03) 726–748.

Huh, Woonghee Tim, Retsef Levi, Paat Rusmevichientong, James B Orlin. 2011. Adaptive data-driven inventory control with censored demand based on kaplan-meier estimator. *Operations Research* **59**(4) 929–941.

Imbens, Guido W, Donald B Rubin. 2015. *Causal inference in statistics, social, and biomedical sciences*. Cambridge University Press.

Kakade, Sham, Karthik Sridharan, Ambuj Tewari. 2008. On the complexity of linear prediction: Risk bounds, margin bounds, and regularization. *NIPS*. 793–800.

Kallus, Nathan. 2014. Predicting crowd behavior with big public data. *WWW*. 23, 625630.

Kao, Yi-hao, Benjamin V Roy, Xiang Yan. 2009. Directed regression. *Advances in Neural Information Processing Systems*. 889–897.

Kaplan, Edward L, Paul Meier. 1958. Nonparametric estimation from incomplete observations. *Journal of the American statistical association* **53**(282) 457–481.

Kleywegt, Anton, Alexander Shapiro, Tito Homem-de Mello. 2002. The sample average approximation method for stochastic discrete optimization. *SIAM J. Optim.* **12**(2) 479–502.

Koenker, Roger. 2005. *Quantile regression*. 38, Cambridge university press.

Lai, Tze Leung, Herbert Robbins. 1985. Asymptotically efficient adaptive allocation rules. *Advances in applied mathematics* **6**(1) 4–22.

Ledoux, Michel, Michel Talagrand. 1991. *Probability in Banach Spaces: isoperimetry and processes*. Springer.

Lehmann, Erich Leo, George Casella. 1998. *Theory of point estimation*, vol. 31. Springer.

Mcdonald, Daniel, Cosma Shalizi, Mark Schervish. 2011. Estimating beta-mixing coefficients. *AISTATS*. 516–524.

Mohri, Mehryar, Afshin Rostamizadeh. 2008. Rademacher complexity bounds for non-iid processes. *NIPS*. 1097–1104.

Mohri, Mehryar, Afshin Rostamizadeh, Ameet Talwalkar. 2012. *Foundations of machine learning*. MIT press.

Mokkadem, Abdelkader. 1988. Mixing properties of arma processes. *Stochastic Process. Appl.* **29**(2) 309–315.

Nadaraya, Elizbar. 1964. On estimating regression. *Theory Probab. Appl.* **9**(1) 141–142.

Nemirovski, Arkadi, Anatoli Juditsky, Guanghui Lan, Alexander Shapiro. 2009. Robust stochastic approximation approach to stochastic programming. *SIAM Journal on Optimization* **19**(4) 1574–1609.

Parzen, Emanuel. 1962. On estimation of a probability density function and mode. *The annals of mathematical statistics* 1065–1076.

Robbins, Herbert. 1952. Some aspects of the sequential design of experiments. *Bulletin of the American Mathematical Society* **58** 527–535.

Robbins, Herbert, Sutton Monro. 1951. A stochastic approximation method. *The annals of mathematical statistics* 400–407.

Rosenbaum, Paul R, Donald B Rubin. 1983. The central role of the propensity score in observational studies for causal effects. *Biometrika* **70**(1) 41–55.

Rudin, Cynthia, Gah-Yi Vahn. 2014. The big data newsvendor: Practical insights from machine learning .

Shapiro, Alexander. 2003. Monte carlo sampling methods. *Handbooks Oper. Res. Management Sci.* **10** 353–425.

Shapiro, Alexander, Arkadi Nemirovski. 2005. On complexity of stochastic programming problems. *Continuous optimization*. Springer, 111–146.

Tibshirani, Robert. 1996. Regression shrinkage and selection via the lasso. *Journal of the Royal Statistical Society. Series B (Methodological)* 267–288.

Trevor, Hastie, Tibshirani Robert, Jerome Friedman. 2001. *The Elements of Statistical Learning*, vol. 1. Springer.

Vapnik, Vladimir. 1992. Principles of risk minimization for learning theory. *Advances in neural information processing systems*. 831–838.

Vapnik, Vladimir. 2000. *The nature of statistical learning theory*. springer.

Wald, Abraham. 1949. Statistical decision functions. *The Annals of Mathematical Statistics* 165–205.

Walk, Harro. 2010. Strong laws of large numbers and nonparametric estimation. *Recent Developments in Applied Probability and Statistics*. Springer, 183–214.

Ward, Joe. 1963. Hierarchical grouping to optimize an objective function. *J. Am. Stat. Assoc.* **58**(301) 236–244.

Watson, Geoffrey. 1964. Smooth regression analysis. *Sankhyā A* 359–372.

# Supplement

## 8. Alternative Approaches using Empirical Risk Minimization

In the beginning of Section 2, we noted that the empirical distribution is insufficient for approximating the full-information problem (2). The solution was to consider local neighborhoods in approximating conditional expected costs; these were computed separately for each $x$. Another approach would be to develop an explicit decision rule and impose structure on it. In this section, we consider an approach to constructing a predictive prescription by selecting from a family of linear functions restricted in some norm,

$$\mathcal{F} = \left\{ z(x) = Wx : W \in \mathbb{R}^{d_z \times d_x}, \, ||W|| \leq R \right\}, \tag{25}$$

so to minimize the empirical marginal expected costs as in (4),

$$\hat{z}_N(\cdot) \in \arg\min_{z(\cdot) \in \mathcal{F}} \tfrac{1}{N} \sum_{i=1}^{N} c(z(x^i); y^i).$$

The linear decision rule can be generalized by transforming $X$ to include nonlinear terms or by embedding in a reproducing kernel Hilbert space. We consider two examples of a norm on the matrix of linear coefficients, $W$: the row-wise $p, p'$-norm and the Schatten $p$-norm, which are, respectively,

$$||W|| = ||(\gamma_1 ||W_1||_p, \dots, \gamma_d ||W_d||_p)||_{p'}, \quad ||W|| = \big|\big|(\tau_1, \dots, \tau_{\min\{d_z, d_x\}})\big|\big|_p,$$

where $\tau_i$ are $W$'s singular values. For example, the Schatten 1-norm is the matrix nuclear norm. In either case, the restriction on the norm is equivalent to an appropriately-weighted regularization term incorporated into the objectives of (4).

Problem (4) corresponds to the traditional framework of empirical risk minimization in statistical learning with a general loss function. It is also closely related to $M$-estimation Geer (2000), except that we are concerned with out-of-sample performance rather than inference, an infinite-dimensional decision rule rather than a finite-dimensional parameter, and potentially non-smooth functions. For $d_z = d_y = 1$, $\mathcal{Z} = \mathbb{R}$, and $c(z; y) = (z - y)^2$, problem (4) corresponds to least-squares regression. For $d_z = d_y = 1$, $\mathcal{Z} = \mathbb{R}$, and $c(z; y) = (y - z)(\tau - \mathbb{I}[y - z < 0])$, problem (4) corresponds to quantile regression (cf. Koenker (2005)), which estimates the conditional $\tau$-quantile as a function of $x$. Rearranging terms, $c(z; y) = (y - z)(\tau - \mathbb{I}[y - z < 0]) = \max\{(1 - \tau)(z - y), \, \tau(y - z)\}$ is the same as the newsvendor cost function where $\tau$ is the service level requirement as observed by Rudin and Vahn (2014). Standard ERM generalization theory deals only with univariate-valued functions. Because most OR/MS problems involve multivariate uncertainty and decisions, in this section we generalize the approach and its associated theoretical guarantees to such multivariate problems ($d_y \geq 1$, $d_z \geq 1$). In particular, we generalize Rademacher complexity to multivariate-valued decision rules and extend the Rademacher comparison Lemma (Theorem 4.12 of Ledoux and Talagrand (1991)) to this new definition. We can then apply standard results to obtain out-of-sample guarantees.

Before continuing, we note a few limitations of any approach based on (4). For general problems, there is no reason to expect that optimal solutions will have a linear structure (whereas certain distributional assumptions lead to such conclusions in least-squares and quantile regression analyses). In particular, unlike the predictive prescriptions studied in Section 2, the approach based on (4) does not enjoy the same universal guarantees of asymptotic optimality. Instead, we will only have out-of-sample guarantees that depend on our class $\mathcal{F}$ of decision rules.

Another limitation is the difficulty in restricting the decisions to a constrained feasible set $\mathcal{Z} \neq \mathbb{R}^{d_z}$. Consider, for example, the portfolio allocation problem from Section 1.1, where we must have $\sum_{i=1}^{d_x} z_i = 1$. One approach to applying (4) to this problem might be to set $c(z; y) = \infty$ for $z \notin \mathcal{Z}$ (or, equivalently, constrain $z(x^i) \in \mathcal{Z} \; \forall i$). However, not only will this not guarantee that $z(x) \in \mathcal{Z}$ for $x$ outside the dataset, but we would also run into a problem of infeasibility as we would have $N$ linear equality constraints on $d_z \times d_x$ linear coefficients (a constraint such as $\sum_{i=1}^{d_x} z_i \leq 1$ that does not reduce the affine dimension will still lead to an undesirably flat linear decision rule as $N$ grows). Another approach may be to compose $\mathcal{F}$ with a projection onto $\mathcal{Z}$, but this will generally lead to a non-convex optimization problem that is intractable to solve. Therefore, the approach is limited in its applicability to OR/MS problems.

In a few limited cases, we may be able to sensibly extend the cost function synthetically outside the feasible region while maintaining convexity. For example, in the shipment planning example of Section 1.1, we may allow negative order quantities $z$ and extend the first-stage costs to depend only on the positive part of $z$, i.e. $p_1 \sum_{i=1}^{d_z} \max\{z_i, 0\}$ (but leave the second-stage costs as they are for convexity). Now, if after training $\hat{z}_N(\cdot)$, we transform any resulting decision by only taking the positive part of each order quantity, we end up with a feasible decision rule whose costs are no worse than the synthetic costs of the original rule.

In the rest of this section we consider the application of the approach (4) to problems where $y$ and $z$ are multivariate and $c(z; y)$ is general, but only treat unconstrained decisions $\mathcal{Z} = \mathbb{R}^{d_z}$.

## 8.1. Tractability

We first develop sufficient conditions for the problem (4) to be optimized in polynomial time. The proof appears in Section 11.

THEOREM 12. *Suppose that $c(z; y)$ is convex in $z$ for every fixed $y$ and let oracles be given for evaluation and subgradient in $z$. Then for any fixed $x$ we can find an $\epsilon$-optimal solution to (4) in time and oracle calls polynomial in $n$, $d$, $\log(1/\epsilon)$ for $\mathcal{F}$ as in (25).*

## 8.2. Out-of-Sample Guarantees

Next, we characterize the out-of-sample guarantees of a predictive prescription derived from (4). All proofs are in the E-companion. In the traditional framework of empirical risk minimization in statistical learning such guarantees are often derived using Rademacher complexity but these only apply to univariate problems (c.f. Bartlett and Mendelson (2003)). Because most OR/MS problems are multivariate, we generalize this theory appropriately. We begin by generalizing the definition of Rademacher complexity to multivariate-valued functions.

DEFINITION 3. Given a sample $S_N = \{s_1, \ldots, s_N\}$, The *empirical multivariate Rademacher complexity* of a class of functions $\mathcal{F}$ taking values in $\mathbb{R}^d$ is defined as

$$\widehat{\mathfrak{R}}_N(\mathcal{F}; S_N) = \mathbb{E}\left[\frac{2}{N}\sup_{g\in\mathcal{F}}\sum_{i=1}^{n}\sum_{k=1}^{d}\sigma_{ik}g_k(s_i)\Big|s_1, \ldots, s_n\right]$$

where $\sigma_{ik}$ are independently equiprobably $+1, -1$. The *marginal multivariate Rademacher complexity* is defined as the expectation over the sampling distribution of $S_N$: $\mathfrak{R}_N(\mathcal{F}) = \mathbb{E}\left[\widehat{\mathfrak{R}}_n(\mathcal{F}; S_N)\right]$.

Note that given only data $S_N$, the quantity $\widehat{\mathfrak{R}}_N(\mathcal{F}; S_N)$ is observable. Note also that when $d = 1$ the above definition coincides with the common definition of Rademacher complexity.

The theorem below relates the multivariate Rademacher complexity of $\mathcal{F}$ to out-of-sample guarantees on the performance of the corresponding predictive prescription $\hat{z}_N(x)$ from (4). A generalization of the following to mixing processes is given in the supplemental Section 10. We denote by $S_N^x = \{x^1, \ldots, x^N\}$ the restriction of our sample to data on $X$.

THEOREM 13. *Suppose $c(z; y)$ is bounded and equi-Lipschitz in $z$:*

$$\sup_{z\in\mathcal{Z}, y\in\mathcal{Y}} c(z; y) \leq \bar{c}, \quad \sup_{z\neq z'\in\mathcal{Z}, y\in\mathcal{Y}} \frac{c(z;y)-c(z';y)}{||z_k-z_k'||_\infty} \leq L < \infty.$$

*Then, for any $\delta > 0$, each of the following events occurs with probability at least $1 - \delta$,*

$$\mathbb{E}\left[c(z(X); Y)\right] \leq \frac{1}{N}\sum_{i=1}^{N}c(z(x^i); y^i) + \bar{c}\sqrt{\log(1/\delta')/2N} + L\mathfrak{R}_N(\mathcal{F}) \qquad \forall z\in\mathcal{F}, \qquad (26)$$

$$\mathbb{E}\left[c(z(X); Y)\right] \leq \frac{1}{N}\sum_{i=1}^{N}c(z(x^i); y^i) + 3\bar{c}\sqrt{\log(2/\delta'')/2N} + L\widehat{\mathfrak{R}}_N(\mathcal{F}; S_N^x) \qquad \forall z\in\mathcal{F}. \qquad (27)$$

*In particular, these hold for $z = \hat{z}_N(\cdot) \in \mathcal{F}$.*

Equations (26) and (27) provide a bound on the out-of-sample performance of any predictive prescription $z(\cdot) \in \mathcal{F}$. The bound is exactly what we minimize in problem (4) because the extra terms do not depend on $z(\cdot)$. That is, we minimize the empirical risk, which, with additional confidence terms, bounds the true out-of-sample costs of the resulting predictive prescription $\hat{z}_N(\cdot)$.

To prove Theorem 13, we first establish a comparison lemma that is an extension of Theorem 4.12 of Ledoux and Talagrand (1991) to our multivariate case.

LEMMA 1. *Suppose that $c$ is $L$-Lipschitz uniformly over $y$ with respect to $\infty$-norm:*

$$\sup_{z\neq z'\in\mathcal{Z}, y\in\mathcal{Y}} \frac{c(z;y)-c(z';y)}{\max_{k=1,\ldots,d}|z_k-z_k'|} \leq L < \infty.$$

Let $\mathcal{G} = \{(x, y) \mapsto c(f(x); y) : f \in \mathcal{F}\}$. Then we have that $\widehat{\mathfrak{R}}_n(\mathcal{G}; S_N) \leq L\widehat{\mathfrak{R}}_n(\mathcal{F}; S_N^x)$ and therefore also that $\mathfrak{R}_n(\mathcal{G}) \leq L\mathfrak{R}_n(\mathcal{F})$. (Notice that one is a univariate complexity and one multivariate and that the complexity of $\mathcal{F}$ involves only the sampling of $x$.)

*Proof*    Write $\phi_i(z) = c(z; y^i)/L$. Then by Lipschitz assumption and by part 2 of Proposition 2.2.1 from Bertsekas et al. (2003), for each $i$, $\phi_i$ is 1-Lipschitz. We now would like to show the inequality in

$$\widehat{\mathfrak{R}}_n(\mathcal{G}; S_N) = \mathbb{E}\left[ \frac{2}{n} \sup_{z \in \mathcal{F}} \sum_{i=1}^n \sigma_{i0} \phi_i(z(x^i)) \,\middle|\, S_N \right]$$
$$\leq L\mathbb{E}\left[ \frac{2}{n} \sup_{z \in \mathcal{F}} \sum_{i=1}^n \sum_{k=1}^d \sigma_{ik} z_k(x^i) \,\middle|\, S_N^x \right]$$
$$= L\widehat{\mathfrak{R}}_n(\mathcal{F}; S_N^x).$$

By conditioning and iterating, it suffices to show that for any $T \subset \mathbb{R} \times \mathcal{Z}$ and 1-Lipschitz $\phi$,

$$\mathbb{E}\left[ \sup_{t,z \in T} (t + \sigma_0 \phi(z)) \right] \leq \mathbb{E}\left[ \sup_{t,z \in T} \left( t + \sum_{k=1}^d \sigma_k z_k \right) \right]. \tag{28}$$

The expectation on the left-hand-side is over two values ($\sigma_0 = \pm 1$) so there are two choices of $(t, z)$, one for each scenario. Let any $(t^{(+1)}, z^{(+1)}), (t^{(-1)}, z^{(-1)}) \in T$ be given. Let $k^*$ and $s^* = \pm 1$ be such that

$$\max_{k=1,\dots,d} \left| z_k^{(+1)} - z_k^{(-1)} \right| = s^* \left( z_{k^*}^{(+1)} - z_{k^*}^{(-1)} \right).$$

Fix $(\tilde{t}^{(\pm 1)}, \tilde{z}^{(\pm 1)}) = (t^{(\pm s^*)}, z^{(\pm s^*)})$. Then, since these are feasible choices in the inner supremum, choosing $(t, z)(\sigma) = (\tilde{t}^{(\sigma_{k^*})}, \tilde{z}^{(\sigma_{k^*})})$, we see that the right-hand-side of (28) has

$$\text{RHS (28)} \geq \frac{1}{2} \mathbb{E}\left[ \tilde{t}^{(+1)} + \tilde{z}_{k^*}^{(+1)} + \sum_{k \neq k^*} \sigma_k \tilde{z}_k^{(+1)} \right]$$
$$+ \frac{1}{2} \mathbb{E}\left[ \tilde{t}^{(-1)} - \tilde{z}_{k^*}^{(-1)} + \sum_{k \neq k^*} \sigma_k \tilde{z}_k^{(-1)} \right]$$
$$= \frac{1}{2} \left( t^{(+1)} + t^{(-1)} + \max_{k=1,\dots,d} \left| z_k^{(+1)} - z_k^{(-1)} \right| \right)$$
$$\geq \frac{1}{2} \left( t^{(+1)} + \phi\left( z^{(+1)} \right) \right) + \frac{1}{2} \left( t^{(-1)} - \phi\left( z^{(-1)} \right) \right)$$

where the last inequality is due to the Lipschitz condition. Since true for any $(t^{(\pm 1)}, z^{(\pm 1)})$ given, taking suprema over the left-hand-side completes the proof.    $\square$

Next, we restate the main result of Bartlett and Mendelson (2003):

THEOREM 14.    *Consider a class $\mathcal{G}$ of functions $\mathcal{U} \to \mathbb{R}$ that are bounded: $|g(u)| \leq \overline{g}$ $\forall g \in \mathcal{G}$, $u \in \mathcal{U}$. Consider a sample $S_n = (u^1, \dots, u^N)$ of some random variable $T \in \mathcal{T}$. Fix $\delta > 0$. Then we have that, with probability $1 - \delta$,*

$$\mathbb{E}\left[ g(T) \right] \leq \frac{1}{N} \sum_{i=1}^N g(u^i) + \overline{g}\sqrt{\log(1/\delta)/2N} + \mathfrak{R}_N(\mathcal{G}) \qquad\qquad \forall g \in \mathcal{G}, \tag{29}$$

*and that, again with probability $1 - \delta$,*

$$\mathbb{E}\left[g(T)\right] \leq \frac{1}{N}\sum_{i=1}^{N} g(u^i) + 3\bar{g}\sqrt{\log(2/\delta)/2N} + \widehat{\mathfrak{R}}_N(\mathcal{G}) \qquad \forall g \in \mathcal{G}. \tag{30}$$

Finally, we can prove Theorem 13:

*Proof of Theorem 13*  Apply Theorem 14 to the random variable $U = (X, Y)$ and function class $\mathcal{G} = \{(x, y) \mapsto c(f(x); y) : f \in \mathcal{F}\}$. Note that by assumption we have boundedness of functions in $\mathcal{G}$ by the constant $\bar{c}$. Bound the complexity of $\mathcal{G}$ by that of $\mathcal{F}$ using Lemma 1 and the assumption of $c(z; y)$ being $L$-Lipschitz. Equations (32) and (33) hold for every $g \in \mathcal{G}$ and hence for every $f \in \mathcal{F}$ and $g(x, y) = c(f(x); y)$, of which the expectation is the expected costs of the decision rule $f$. $\quad\square$

Equations (26) and (27) in Theorem 13 involve the multivariate Rademacher complexity of our class $\mathcal{F}$ of decision rules. In the next lemmas, we compute appropriate bounds on the complexity of our examples of classes $\mathcal{F}$. The theory, however, applies beyond linear rules.

LEMMA 2.  *Consider $\mathcal{F}$ as in (25) with row-wise $p, p'$ norm for $p \in [2, \infty)$ and $p' \in [1, \infty]$. Let $q$ be the conjugate exponent of $p$ $(1/p + 1/q = 1)$ and suppose that $\lVert x \rVert_q \leq M$ for all $x \in \mathcal{X}$. Then*

$$\mathfrak{R}_N(\mathcal{F}) \leq 2MR\sqrt{\frac{p-1}{N}}\sum_{k=1}^{d_z}\frac{1}{\gamma_k}.$$

LEMMA 3.  *Consider $\mathcal{F}$ as in (25) with Schatten $p$-norm. Let $r = \max\{1 - 1/p, 1/2\}$. Then*

$$\widehat{\mathfrak{R}}_N(\mathcal{F}; S_N^x) \leq 2Rd_z^r\sqrt{\frac{1}{N}}\sqrt{\frac{1}{N}\sum_{i=1}^{N}\lVert x^i \rVert}, \quad \mathfrak{R}_N(\mathcal{F}) \leq 2Rd_z^r\sqrt{\frac{1}{N}}\sqrt{\mathbb{E}\lVert X \rVert_2^2}.$$

The above results indicate that the confidence terms in equations (26) and (27) shrink to 0 as $N \to \infty$ even if we slowly relax norm restrictions. Hence, we can approach the optimal out-of-sample performance over the class $\mathcal{F}$ without restrictions on norms.

*Proof of Lemma 2*  Consider $\mathcal{F}_k = \{z_k(\cdot) : z \in \mathcal{F}\} = \{z_k(x) = w^T x : \lVert w \rVert_p \leq \frac{R}{\gamma_k}\}$, the projection of $\mathcal{F}$ onto the $k^{\text{th}}$ coordinate. Then $\mathcal{F} \subset \mathcal{F}_1 \times \cdots \times \mathcal{F}_{d_z}$ and $\mathfrak{R}_N(\mathcal{F}) \leq \sum_{k=1}^{d_z}\mathfrak{R}_N(\mathcal{F}_k)$. The latter right-hand-side complexities are the common univariate Rademacher complexities. Applying Theorem 1 of Kakade et al. (2008) to each component we get $\mathfrak{R}_N(\mathcal{F}_k) \leq 2M\sqrt{\frac{p-1}{N}}\frac{R}{\gamma_k}$. $\quad\square$

*Proof of Lemma 3*  Let $q$ be $p$'s conjugate exponent $(1/p + 1/q = 1)$. In terms of vector norms on $v \in \mathbb{R}^d$, if $q \geq 2$ then $\lVert v \rVert_p \leq \lVert v \rVert_2$ and if $q \leq 2$ then $\lVert b \rVert_p \leq d^{1/2 - 1/p}\lVert v \rVert_2$. Let $F$ be the matrix $F_{ji} = x_j^i$. Note that $F\sigma \in \mathbb{R}^{d_x \times d_z}$. By Jensen's inequality and since Schatten norms are vector norms on singular values,

$$\begin{aligned}
\widehat{\mathfrak{R}}_N^2(\mathcal{F}; S_N^x) &\leq \frac{4}{N^2}\mathbb{E}\left[\sup_{\lVert W \rVert_p \leq R}\text{Trace}\left(WF\sigma\right)^2 \Big| S_N^x\right] \\
&= \frac{4R^2}{N^2}\mathbb{E}\left[\lVert F\sigma \rVert_q^2 \big| S_N^x\right] \\
&\leq \frac{4R^2}{N^2}\max\left\{\min\{d_z, d_x\}^{1-2/p}, 1\right\}\mathbb{E}\left[\lVert F\sigma \rVert_2^2 \big| S_N^x\right] \\
&\leq \frac{4R^2}{N^2}\max\left\{d_z^{1-2/p}, 1\right\}\mathbb{E}\left[\lVert F\sigma \rVert_2^2 \big| S_N^x\right].
\end{aligned}$$

The first result follows because

$$\frac{1}{N}\mathbb{E}\left[\left\|F\sigma\right\|_2^2 \big| S_N^x\right] = \frac{1}{n}\sum_{k=1}^{d_z}\sum_{j=1}^{d_x}\sum_{i,i'=1}^{N} x_j^i x_j^{i'}\mathbb{E}\left[\sigma_{ik}\sigma_{i'k}\right]$$

$$= \frac{d_z}{N}\sum_{i=1}^{N}\sum_{j=1}^{d_x}(x_j^i)^2 = d_z\,\widehat{\mathbb{E}}_N\left\|x\right\|_2^2$$

The second result follows by applying Jensen's inequality again to pass the expectation over $S_n$ into the square. $\qquad\square$

## 9. Extensions of Asymptotic Optimality to Mixing Processes and Proofs

In this supplemental section, we generalize the asymptotic results to mixing process and provide the omitted proofs from Sections 4.3 and 4.4.

### 9.1. Mixing Processes

We begin by defining stationary and mixing processes.

DEFINITION 4. A sequence of random variables $V_1, V_2, \dots$ with joint measure $\mu$ is called *stationary* if joint distributions of finitely many consecutive variables are invariant to shifting. That is,

$$\mu_{V_t,\dots,V_{t+k}} = \mu_{V_s,\dots,V_{s+k}} \quad \forall s,t \in \mathbb{N},\, k \geq 0,$$

where $\mu_{V_t,\dots,V_{t+k}}$ is the induced measure on a sequence of length $k$.

In particular, if a sequence is stationary then the variables have identical marginal distributions, but they may not be independent and the sequence may not be exchangeable. Instead of independence, mixing is the property that if standing at particular point in the sequence we look far enough ahead, the head and the tail look nearly independent, where "nearly" is defined by different metrics for different definitions of mixing.

DEFINITION 5. Given a stationary sequence $\{V_t\}_{t\in\mathbb{N}}$, denote by $\mathcal{A}^t = \sigma(V_1,\dots,V_t)$ the sigma-algebra generated by the first $t$ variables and by $\mathcal{A}_t = \sigma(V_t, V_{t+1}, \dots)$ the sigma-algebra generated by the subsequence starting at $t$. Define the *mixing coefficients at lag $k$*

$$\alpha(k) = \sup_{t\in\mathbb{N},\, A\in\mathcal{A}^t,\, B\in\mathcal{A}_{t+k}} |\mu(A \cap B) - \mu(A)\mu(B)|$$

$$\beta(k) = \sup_{t\in\mathbb{N}} \left\|\mu_{\{V_s\}_{s\leq t}} \otimes \mu_{\{V_s\}_{s\geq t+k}} - \mu_{\{V_s\}_{s\leq t \vee s\geq t+k}}\right\|_{\mathrm{TV}}$$

$$\rho(k) = \sup_{t\in\mathbb{N},\, Q\in L_2(\mathcal{A}^t),\, R\in L_2(\mathcal{A}_{t+k})} |\mathrm{Corr}(Q, R)|$$

where $\left\|\cdot\right\|_{\mathrm{TV}}$ is the total variance and $L_2(\mathcal{A})$ is the set of $\mathcal{A}$-measurable square-integrable real-valued random variables.

$\{V_t\}$ is said to be $\alpha$-mixing if $\alpha(k) \overset{k\to\infty}{\longrightarrow} 0$ , $\beta$-mixing if $\beta(k) \overset{k\to\infty}{\longrightarrow} 0$, and $\rho$-mixing if $\rho(k) \overset{k\to\infty}{\longrightarrow} 0$.

Notice that an iid sequence has $\alpha(k) = \beta(k) = \rho(k) = 0$. Bradley (1986) establishes that $2\alpha(k) \leq \beta(k)$ and $4\alpha(k) \leq \rho(k)$ so that either $\beta$- or $\rho$-mixing implies $\alpha$-mixing.

Many processes satisfy mixing conditions under mild assumptions: auto-egressive moving-average (ARMA) processes (cf. Mokkadem (1988)), generalized autoregressive conditional heteros-kedasticity (GARCH) processes (cf. Carrasco and Chen (2002)), and certain Markov chains. For a thorough discussion and more examples see Doukhan (1994) and Bradley (2005). Mixing rates are often given explicitly by model parameters but they can also be estimated from data (cf. Mcdonald et al. (2011)). Sampling from such processes models many real-life sampling situations where observations are taken from an evolving system such as, for example, the stock market, inter-dependent product demands, or aggregates of doubly stochastic arrival processes as in the posts on social media.

## 9.2. Asymptotic Optimality

Let us now restate the results of Section 4.3 in more general terms, encompassing both iid and mixing conditions on $S_N$.

THEOREM 15 ($k$**NN**). *Suppose Assumptions 3, 4, and 5 hold and that $S_N$ is generated by iid sampling. Let $w_{N,i}(x)$ be as in (12) with $k = \min\left\{\lceil CN^{\delta} \rceil, N-1\right\}$ for some $C > 0, 0 < \delta < 1$. Let $\hat{z}_N(x)$ be as in (3). Then $\hat{z}_N(x)$ is asymptotically optimal and consistent.*

THEOREM 16 (**Kernel Methods**). *Suppose Assumptions 3, 4, and 5 hold and that $\mathbb{E}\left[|c(z;Y)|\max\left\{\log|c(z;Y)|, 0\right\}\right] < \infty$ for each $z$. Let $w_{N,i}(x)$ be as in (13) with $K$ being any of the kernels in Section 2.2 and $h = CN^{-\delta}$ for $C, \delta > 0$. Let $\hat{z}_N(x)$ be as in (3). If $S_N$ comes from*

1. *an iid process and $\delta < 1/d_x$, or*

2. *a $\rho$-mixing process with $\rho(k) = O(k^{-\gamma})$ $(\gamma > 0)$ and $\delta < 2\gamma/(d_x + 2d_x\gamma)$, or*

3. *an $\alpha$-mixing process with $\alpha(k) = O(k^{-\gamma})$ $(\gamma > 1)$ and $\delta < 2(\gamma-1)/(3d_x + 2d_x\gamma)$,*

*then $\hat{z}_N(x)$ is asymptotically optimal and consistent.*

THEOREM 17 (**Recursive Kernel Methods**). *Suppose Assumptions 3, 4, and 5 hold and that $S_N$ comes from a $\rho$-mixing process with $\sum_{k=1}^{\infty} \rho(k) < \infty$ (or iid). Let $w_{N,i}(x)$ be as in (14) with $K$ being the naïve kernel and with $h_i = Ci^{-\delta}$ for some $C > 0$, $0 < \delta < 1/(2d_x)$. Let $\hat{z}_N(x)$ be as in (3). Then $\hat{z}_N(x)$ is asymptotically optimal and consistent.*

THEOREM 18 (**Local Linear Methods**). *Suppose Assumptions 3, 4, and 5 hold, that $\mu_X$ is absolutely continuous and has density bounded away from 0 and $\infty$ on the support of $X$ and twice continuously differentiable, and that costs are bounded over $y$ for each $z$ (i.e., $|c(z;y)| \leq g(z)$) and twice continuously differentiable. Let $w_{N,i}(x)$ be as in (15) with $K$ being any of the kernels in Section 2.2 and with $h_N = CN^{-\delta}$ for some $C, \delta > 0$. Let $\hat{z}_N(x)$ be as in (3). If $S_N$ comes from*

1. *an iid process and $\delta < 1/d_x$, or*

2. *an $\alpha$-mixing process with $\alpha(k) = O(k^{-\gamma})$, $\gamma > d_x + 3$, and $\delta < (\gamma - d_x - 3)/(d_x(\gamma - d_x + 3))$,*

*then $\hat{z}_N(x)$ is asymptotically optimal and consistent.*

THEOREM 19 (**Nonnegative Local Linear Methods**). *Suppose Assumptions 3, 4, and 5 hold, that $\mu_X$ is absolutely continuous and has density bounded away from 0 and $\infty$ on the support of $X$ and twice continuously differentiable, and that costs are bounded over $y$ for each $z$ (i.e., $|c(z; y)| \le g(z)$) and twice continuously differentiable. Let $w_{N,i}(x)$ be as in (16) with $K$ being any of the kernels in Section 2.2 and with $h_N = C N^{-\delta}$ for some $C, \delta > 0$. Let $\hat{z}_N(x)$ be as in (3). If $S_N$ comes from*

1. *an iid process and $\delta < 1/d_x$, or*

2. *an $\alpha$-mixing process with $\alpha(k) = O(k^{-\gamma})$, $\gamma > d_x + 3$, and $\delta < (\gamma - d_x - 3)/(d_x(\gamma - d_x + 3))$,*

*then $\hat{z}_N(x)$ is asymptotically optimal and consistent.*

THEOREM 20 (**Decisions Affect Uncertainty**). *Suppose Assumptions 1, 2, 3, 4, and 5 (case 1) hold, that $\mu_{(X,Z_1)}$ is absolutely continuous and has density bounded away from 0 and $\infty$ on the support of $X, Z_1$ and twice continuously differentiable, and that costs are bounded over $y$ for each $z$ (i.e., $|c(z; y)| \le g(z)$) and twice continuously differentiable. Let $w_{N,i}(\tilde{x})$ be as in (13), (15), or (16) applied to $\tilde{S}_N$ with $K$ being any of the kernels in Section 2.2 and with $h_N = C N^{-\delta}$ for some $C > 0$, $\delta > 0$. Let $\hat{z}_N(x)$ be as in (21). If $\tilde{S}_N$ comes from*

1. *an iid process and $\delta < 1/(d_x + d_{z_1})$, or*

2. *an $\alpha$-mixing process with $\alpha(k) = O(k^{-\gamma})$, $\gamma > d_x + d_{z_1} + 3$, and $\delta < (\gamma - d_x - d_{z_1} - 3)/(d_x(\gamma - d_x - d_{z_1} + 3))$,*

*Then $\hat{z}_N(x)$ is asymptotically optimal and consistent.*

### 9.3. Proofs of Asymptotic Results for Local Predictive Prescriptions

First, we establish some preliminary results. In what follows, let

$$C(z|x) = \mathbb{E}\left[c(z; Y) \,\big|\, X = x\right],$$

$$\widehat{C}_N(z|x) = \sum_{i=1}^{N} w_{N,i}(x) c(z; y^i),$$

$$\mu_{Y|x}(A) = \mathbb{E}\left[\mathbb{I}\left[Y \in A\right] \,\big|\, X = x\right],$$

$$\hat{\mu}_{Y|x,N}(A) = \sum_{i=1}^{N} w_{N,i}(x) \mathbb{I}\left[y^i \in A\right].$$

LEMMA 4. *If $\{(x^i, y^i)\}_{i \in \mathbb{N}}$ is stationary and $f : \mathbb{R}^{m_Y} \to \mathbb{R}$ is measurable then $\{(x^i, f(y^i))\}_{i \in \mathbb{N}}$ is also stationary and has mixing coefficients no larger than those of $\{(x^i, y^i)\}_{i \in \mathbb{N}}$.*

*Proof* This is simply because a transform can only make the generated sigma-algebra coarser. For a single time point, if $f$ is measurable and $B \in \mathcal{B}(\mathbb{R})$ then by definition $f^{-1}(B) \in \mathcal{B}(\mathbb{R})$ and, therefore, $\{Y^{-1}(f^{-1}(B)) : B \in \mathcal{B}(\mathbb{R})\} \subset \{Y^{-1}(B) : B \in \mathcal{B}(\mathbb{R}^{m_Y})\}$. Here the transform is applied independently across time so the result holds ($f \times \cdots \times f$ remains measurable). $\qquad \square$

**Lemma 5.** *Suppose Assumptions 3 and 4 hold. Fix $x \in \mathcal{X}$ and a sample path of data such that, for every $z \in \mathcal{Z}$, $\widehat{C}_N(z|x) \to C(z|x)$. Then $\widehat{C}_N(z|x) \to C(z|x)$ uniformly in $z$ over any compact subset of $\mathcal{Z}$.*

*Proof* Let any convergent sequence $z_N \to z$ and $\epsilon > 0$ be given. By equicontinuity and $z_N \to z$, $\exists N_1$ such that $|c(z_N; y) - c(z; y)| \leq \epsilon/2 \ \forall N \geq N_1$. Then $\left|\widehat{C}_N(z_N|x) - \widehat{C}_N(z|x)\right| \leq \mathbb{E}_{\hat{\mu}_{Y|x,N}} |c(z_N; y) - c(z; y)| \leq \epsilon/2 \ \forall N \geq N_1$. By assumption $\widehat{C}_N(z|x) \to C(z|x)$ and hence $\exists N_2$ such that $\left|\widehat{C}_N(z|x) - C(z|x)\right| \leq \epsilon/2$. Therefore, for $N \geq \max\{N_1, N_2\}$,
$$\left|\widehat{C}_N(z_N|x) - C(z|x)\right| \leq \left|\widehat{C}_N(z_N|x) - \widehat{C}_N(z|x)\right| + \left|\widehat{C}_N(z|x) - C(z|x)\right| \leq \epsilon.$$
Hence $\widehat{C}_N(z_N|x) \to C(z|x)$ for any convergent sequence $z_N \to z$.

Now fix $E \subset \mathcal{Z}$ compact and suppose for contradiction that $\sup_{z \in E} \left|\widehat{C}_N(z|x) - C(z|x)\right| \not\to 0$. Then $\exists \epsilon > 0$ and $z_N \in E$ such that $\left|\widehat{C}_N(z_N|x) - C(z_N|x)\right| \geq \epsilon$ infinitely often. Restricting first to a subsequence where this always happens and then using the compactness of $E$, there exists a convergent subsequence $z_{N_k} \to z \in E$ such that $\left|\widehat{C}_{N_k}(z_{N_k}|x) - C(z_{N_k}|x)\right| \geq \epsilon$ for every $k$. Then,
$$0 < \epsilon \leq \left|\widehat{C}_{N_k}(z_{N_k}|x) - C(z_{N_k}|x)\right| \leq \left|\widehat{C}_{N_k}(z_{N_k}|x) - C(z|x)\right| + \left|C(z|x) - C(z_{N_k}|x)\right|.$$
Since $z_{N_k} \to z$, we have shown before that $\exists k_1$ such that $\left|\widehat{C}_{N_k}(z_{N_k}|x) - C(z|x)\right| \leq \epsilon/2 \ \forall k \geq k_1$. By equicontinuity and $z_{N_k} \to z$, $\exists k_2$ such that $|c(z_{N_k}; y) - c(z; y)| \leq \epsilon/4 \ \forall k \geq k_2$. Hence, also $|C(z|x) - C(z_{N_k}|x)| \leq \mathbb{E}\left[|c(z_{N_k}; y) - c(z; y)| \,\big|\, X = x\right] \leq \epsilon/4 \ \forall k \geq k_2$. Considering $k = \max\{k_1, k_2\}$ we get the contradiction that $0 < \epsilon \leq \epsilon/2$. $\qquad \square$

**Lemma 6.** *Suppose Assumptions 3, 4, and 5 hold. Fix $x \in \mathcal{X}$ and a sample path of data such that $\hat{\mu}_{Y|x,N} \to \mu_{Y|x}$ weakly and, for every $z \in \mathcal{Z}$, $\widehat{C}_N(z|x) \to C(z|x)$. Then $\lim_{N \to \infty} \left(\min_{z \in \mathcal{Z}} \widehat{C}_N(z|x)\right) = v^*(x)$ and every sequence $z_N \in \arg\min_{z \in \mathcal{Z}} \widehat{C}_N(z|x)$ satisfies $\lim_{N \to \infty} C(z_N|x) = v^*(x)$ and $\lim_{N \to \infty} \inf_{z \in Z^*(x)} ||z - z_N|| = 0$.*

*Proof* First, we show $\widehat{C}_N(z|x)$ and $C(z|x)$ are continuous and eventually coercive. Let $\epsilon > 0$ be given. By equicontinuity, $\exists \delta > 0$ such that $|c(z; y) - c(z'; y)| \leq \epsilon \ \forall y \in \mathcal{Y}$ whenever $||z - z'|| \leq \delta$. Hence, whenever $||z - z'|| \leq \delta$, we have $\left|\widehat{C}_N(z|x) - \widehat{C}_N(z'|x)\right| \leq \mathbb{E}_{\hat{\mu}_{Y|x,N}} |c(z; y) - c(z'; y)| \leq \epsilon$ and $|C(z|x) - C(z'|x)| \leq \mathbb{E}\left[|c(z; y) - c(z'; y)| \,\big|\, X = x\right] \leq \epsilon$. This gives continuity. Coerciveness is trivial if $\mathcal{Z}$ is bounded. Suppose it is not. Without loss of generality $D_x$ is compact, otherwise we can take any compact subset of it that has positive probability on it. Then by assumption of weak convergence $\exists N_0$ such that $\hat{\mu}_{Y|x,N}(D_x) \geq \mu_{Y|x}(D_x)/2 > 0$ for all $N \geq N_0$. Now let $z_k \in \mathcal{Z}$ be any sequence such that $||z_k|| \to \infty$. Let $M > 0$ be given. Let $\lambda' = \liminf_{k \to \infty} \inf_{y \notin D_x} c(z_k; y)$ and $\lambda = \max\{\lambda', 0\}$. By assumption $\lambda' > -\infty$. Hence $\exists k_0$ such that $\inf_{y \notin D_x} c(z_k; y) \geq \lambda' \ \forall k \geq k_0$. By $D_x$-uniform coerciveness and $||z_k|| \to \infty$, $\exists k_1 \geq k_0$ such that $c(z_k; y) \geq (2M - 2\lambda)/\mu_{Y|x}(D_x) \ \forall k \geq k_1$ and $y \in D_x$. Hence, $\forall k \geq k_1$ and $N \geq N_0$,

$$C(z_k|x) \geq \mu_{Y|x}(D_x) \times (2M - 2\lambda)/\mu_{Y|x}(D_x) + (1 - \mu_{Y|x}(D_x))\lambda' \geq 2M - 2\lambda + \lambda \geq M,$$

$$\widehat{C}_N(z_k|x) \geq \hat{\mu}_{Y|x,N}(D_x) \times (2M - 2\lambda)/\hat{\mu}_{Y|x,N}(D_x) + (1 - \hat{\mu}_{Y|x,N}(D_x))\lambda' \geq M - \lambda + \lambda = M,$$

since $\alpha\lambda' \geq \lambda$ if $\alpha \geq 0$. This gives coerciveness eventually. By the usual extreme value theorem (c.f. Bertsekas (1999), pg. 669), $\widehat{\mathcal{Z}}_N(x) = \arg\min_{z \in \mathcal{Z}} \widehat{C}_N(z|x)$ and $\mathcal{Z}^*(x) = \arg\min_{z \in \mathcal{Z}} C(z|x)$ exist, are nonempty, and are compact.

Now we show there exists $\mathcal{Z}^*_\infty(x)$ compact such that $\mathcal{Z}^*(x) \subset \mathcal{Z}^*_\infty(x)$ and $\widehat{\mathcal{Z}}_N(x) \subset \mathcal{Z}^*_\infty(x)$ eventually. If $\mathcal{Z}$ is bounded this is trivial. So suppose otherwise (and again, without loss of generality $D_x$ is compact). Fix any $z^* \in \mathcal{Z}^*(x)$. Then by Lemma 5 we have $\widehat{C}_N(z^*|x) \to C(z^*|x)$. Since $\min_{z \in \mathcal{Z}} \widehat{C}_N(z|x) \leq \widehat{C}_N(z^*|x)$, we have $\limsup_{N \to \infty} \min_{z \in \mathcal{Z}} \widehat{C}_N(z|x) \leq C(z^*|x) = \min_{z \in \mathcal{Z}} C(z|x) = v^*$. Now suppose for contradiction no such $\mathcal{Z}^*_\infty(x)$ exists. Then there must be a subsequence $z_{N_k} \in \widehat{\mathcal{Z}}_{N_k}$ such that $||z_{N_k}|| \to \infty$. By $D_x$-uniform coerciveness and $||z_{N_k}|| \to \infty$, $\exists k_1 \geq k_0$ such that $c(z_{N_k}; y) \geq 2(v^* + 1 - \lambda)/\mu_{Y|x}(D_x) \ \forall k \geq k_1$ and $y \in D_x$. Hence, $\forall k \geq k_1$ and $N \geq N_0$,

$$\widehat{C}_N(z_{N_k}|x) \geq \hat{\mu}_{Y|x,N}(D_x) \times 2(v^* + 1 - \lambda)/\mu_{Y|x}(D_x) + (1 - \hat{\mu}_{Y|x,N}(D_x)) \geq v^* + 1.$$

This yields a contradiction $v^* + 1 \leq v^*$. So $\mathcal{Z}^*_\infty(x)$ exists.

Applying Lemma 5,

$$\tau_N = \sup_{z \in \mathcal{Z}^*_\infty(x)} \left| \widehat{C}_N(z|x) - C(z|x) \right| \to 0.$$

The first result follows from

$$\delta_N = \left| \min_{z \in \mathcal{Z}} \widehat{C}_N(z|x) - \min_{z \in \mathcal{Z}} C(z|x) \right| \leq \sup_{z \in \mathcal{Z}^*_\infty(x)} \left| \widehat{C}_N(z|x) - C(z|x) \right| = \tau_N \to 0.$$

Now consider any sequence $z_N \in \widehat{\mathcal{Z}}_N(x)$. The second result follows from

$$\left| C(\hat{z}_N|x) - \min_{z \in \mathcal{Z}} C(z|x) \right| \leq \left| \widehat{C}_N(\hat{z}_N(x)|x) - C(\hat{z}_N|x) \right| + \left| \min_{z \in \mathcal{Z}} \widehat{C}_N(z|x) - \min_{z \in \mathcal{Z}} C(z|x) \right| \leq \tau_N + \delta_N \to 0.$$

Suppose the third result is false. Then since $\mathcal{Z}^*_\infty(x)$ is compact, there is a convergent subsequence $z_{N_k} \to z'$ such that $\inf_{z \in \mathcal{Z}^*(x)} ||z_{N_k} - z|| \geq \eta > 0$ for all $k$. Since $\inf_{z \in \mathcal{Z}^*(x)} ||z' - z|| \geq \inf_{z \in \mathcal{Z}^*(x)} ||z_{N_k} - z|| - ||z_{N_k} - z'|| \to \eta > 0$, we have $z' \notin \mathcal{Z}^*(x)$ and hence $\epsilon = C(z'|x) - \min_{z \in \mathcal{Z}} C(z|x) > 0$. By equicontinuity and $z_{N_k} \to z'$, $\exists k_2$ such that $|c(z_{N_k}; y) - c(z'; y)| \leq \epsilon/2 \ \forall y \in \mathcal{Y} \ \forall k \geq k_2$. Then, $|C(z_{N_k}|x) - C(z'|x)| \leq \mathbb{E}\left[ |c(z_{N_k}; y) - c(z'; y)| \ \big| X = x \right] \leq \epsilon/2 \ \forall k \geq k_2$. Therefore, $\forall k \geq k_2$,

$$\tau_{N_k} + \delta_{N_k} \geq C(z_{N_k}|x) - \min_{z \in \mathcal{Z}} C(z|x) \geq C(z'|x) - \min_{z \in \mathcal{Z}} C(z|x) - \epsilon/2 = \epsilon/2,$$

which, taking limits, is a contradiction, yielding the third result. $\qquad\square$

LEMMA 7. *Suppose $c(z; y)$ is equicontinuous in $z$. Suppose moreover that for each fixed $z \in \mathcal{Z} \subset \mathbb{R}^d$ we have that $\widehat{C}_N(z|x) \to C(z|x)$ a.s. for $\mu_X$-a.e.x and that for each fixed measurable $D \subset \mathcal{Y}$ we have that $\hat{\mu}_{Y|x,N}(D) \to \mu_{Y|x}(D)$ a.s. for $\mu_X$-a.e.x. Then, a.s. for $\mu_X$-a.e.x, $\widehat{C}_N(z|x) \to C(z|x)$ for all $z \in \mathcal{Z}$ and $\hat{\mu}_{Y|x,N} \to \mu_{Y|x}$ weakly.*

*Proof* Since Euclidean space is separable, $\hat{\mu}_{Y|x,N} \to \mu_{Y|x}$ weakly a.s. for $\mu_X$-a.e.x (c.f. Theorem 11.4.1 of Dudley (2002)). Consider the set $\mathcal{Z}' = \mathcal{Z} \cap \mathbb{Q}^d \cup \{\text{the isolated points of } \mathcal{Z}\}$. Then $\mathcal{Z}'$ is countable and dense in $\mathcal{Z}$. Since $\mathcal{Z}'$ is countable, by continuity of measure, a.s. for $\mu_X$-a.e.x, $\widehat{C}_N(z'|x) \to C(z'|x)$ for all $z' \in \mathcal{Z}'$. Restrict to a sample path and $x$ where this event occurs. Consider any $z \in \mathcal{Z}$ and $\epsilon > 0$. By equicontinuity $\exists \delta > 0$ such that $|c(z; y) - c(z'; y)| \le \epsilon/2$ whenever $||z - z'|| \le \delta$. By density there exists such $z' \in \mathcal{Z}'$. Then, $\left| \widehat{C}_N(z|x) - \widehat{C}_N(z'|x) \right| \le \mathbb{E}_{\hat{\mu}_{Y|x,N}} \left[ |c(z; y) - c(z'; y)| \right] \le \epsilon/2$ and $|C(z|x) - C(z'|x)| \le \mathbb{E}\left[ |c(z; y) - c(z'; y)| \, \big| \, X = x \right] \le \epsilon/2$. Therefore, $0 \le \left| \widehat{C}_N(z|x) - C(z|x) \right| \le \left| \widehat{C}_N(z|x) - C(z'|x) \right| + \epsilon \to \epsilon$. Since true for each $\epsilon$, the result follows for all $z \in \mathcal{Z}$. The choice of particular sample path and $x$ constitute a measure-1 event by assumption. $\qquad\square$

Now, we prove the general form of the asymptotic results from Section 9.2.

*Proof of Theorem 15* Fix $z \in \mathcal{Z}$. Set $Y' = c(z; y)$. By Assumption 3, $\mathbb{E}[|Y'|] < \infty$. Let us apply Theorem 5 of Walk (2010) to $Y'$. By iid sampling and choice of $k$, we have that $\widehat{C}_N(z|x) \to \mathbb{E}[Y'|X = x]$ for $\mu_X$-a.e.x, a.s.

Now fix $D$ measurable. Set $Y' = \mathbb{I}[y \in D]$. Then $\mathbb{E}[Y']$ exists by measurability and $Y'$ is bounded in $[0, 1]$. Therefore applying Theorem 5 of Walk (2010) in the same manner again, $\hat{\mu}_{Y|x,N}(D)$ converges to $\mu_{Y|x}(D)$ for $\mu_X$-a.e.x a.s.

Applying Lemma 7 we obtain that assumptions for Lemma 6 hold for $\mu_X$-a.e.x, a.s., which in turn yields the result desired. $\qquad\square$

*Proof of Theorem 16* Fix $z \in \mathcal{Z}$. Set $Y' = c(z; y)$. By Assumption 3, $\mathbb{E}[|Y'|] < \infty$. Let us apply Theorem 3 of Walk (2010) to $Y'$. By assumption in theorem statement, we also have that $\mathbb{E}\left\{ |Y'| \max\left\{ \log|Y'|, 0 \right\} \right\} < \infty$. Moreover each of the kernels in Section 2.2 can be rewritten $K(x) = H(||x||)$ such that $H(0) > 0$ and $\lim_{t \to \infty} t^d x H(t) \to 0$.

Consider the case of iid sampling. Then our data on $(X, Y')$ is $\rho$-mixing with $\rho(k) = 0$. Using these conditions and our choices of kernel and $h_N$, Theorem 3 of Walk (2010) gives that $\widehat{C}_N(z|x) \to \mathbb{E}[Y'|X = x]$ for $\mu_X$-a.e.x, a.s.

Consider the case of $\rho$-mixing or $\alpha$-mixing. By Lemma 4, equal or lower mixing coefficients hold for $X, Y'$ as hold for $X, Y$. Using these conditions and our choices of kernel and $h_N$, Theorem 3 of Walk (2010) gives that $\widehat{C}_N(z|x) \to \mathbb{E}[Y'|X = x]$ for $\mu_X$-a.e.x, a.s.

Now fix $D$ measurable. Set $Y' = \mathbb{I}\,[y \in D]$. Then $\mathbb{E}[Y']$ exists by measurability and $\mathbb{E}\{|Y'|\max\{\log|Y'|, 0\}\} \le 1 < \infty$. Therefore applying Theorem 3 of Walk (2010) in the same manner again, $\hat{\mu}_{Y|x,N}(D)$ converges to $\mu_{Y|x}(D)$ for $\mu_X$-a.e.x a.s.

Applying Lemma 7 we obtain that assumptions for Lemma 6 hold for $\mu_X$-a.e.x, a.s., which in turn yields the result desired. $\qquad\square$

*Proof of Theorem 17* Fix $z \in \mathcal{Z}$. Set $Y' = c(z; y)$. By Assumption 3, $\mathbb{E}[|Y'|] < \infty$. Let us apply Theorem 4 of Walk (2010) to $Y'$. Note that the naïve kernel satisfies the necessary conditions.

Since our data on $(X, Y)$ is $\rho$-mixing by assumption, we have that by Lemma 4, equal or lower mixing coefficients hold for $X, Y'$ as hold for $X, Y$. Using these conditions and our choice of the naïve kernel and $h_N$, Theorem 4 of Walk (2010) gives that $\widehat{C}_N(z|x) \to \mathbb{E}[Y'|X = x]$ for $\mu_X$-a.e.x, a.s.

Now fix $D$ measurable. Set $Y' = \mathbb{I}\,[y \in D]$. Then $\mathbb{E}[Y']$ exists by measurability. Therefore applying Theorem 4 of Walk (2010) in the same manner again, $\hat{\mu}_{Y|x,N}(D)$ converges to $\mu_{Y|x}(D)$ for $\mu_X$-a.e.x a.s.

Applying Lemma 7 we obtain that assumptions for Lemma 6 hold for $\mu_X$-a.e.x, a.s., which in turn yields the result desired. $\qquad\square$

*Proof of Theorem 18* Fix $z \in \mathcal{Z}$ and $x \in \mathcal{X}$. Set $Y' = c(z; Y)$. By Assumption 3, $\mathbb{E}[|Y'|] < \infty$. Let us apply Theorem 11 of Hansen (2008) to $Y'$ and use the notation thereof. Fix the neighborhood of consideration to the point $x$ (i.e., set $c_N = 0$) since uniformity in $x$ is not of interest. All of the kernels in Section 2.2 are bounded above and square integrable and therefore satisfy Assumption 1 of Hansen (2008). Let $f$ be the density of $X$. By assumption $0 < \delta \le f(x) \le B_0 < \infty$ for all $x \in \mathcal{X}$. Moreover, our choice of $h_N$ satisfies $h_N \to 0$.

Consider first the iid case. Then we have $\alpha(k) = 0 = O(k^{-\gamma})$ for $\gamma = \infty$ ($\beta$ in Hansen (2008)). Combined with boundedness conditions of $Y'$ and $f$ ($|Y'| \le g(z) < \infty$ and $\delta < f < B_0$), we satisfy Assumption 2 of Hansen (2008). Setting $\gamma = \infty$, $s = \infty$ in (17) of Hansen (2008) we get $\theta = 1$. Therefore, since $h = O(N^{-1/d_x})$ we have

$$\frac{(\log\log N)^4 (\log N)^2}{N^\theta h_N^{d_x}} \to 0.$$

Having satisfied all the conditions of Theorem 11 of Hansen (2008), we have that $\widehat{C}_N(z|x) \to \mathbb{E}[Y'|X = x]$ a.s.

Now consider the $\alpha$-mixing case. If the mixing conditions hold for $X, Y$ then by Lemma 4, equal or lower mixing coefficients hold for $X, Y'$. By letting $s = \infty$ we have $\gamma > d_x + 3 > 2$. Combined with boundedness conditions of $Y'$ and $f$ ($|Y'| \le g(z) < \infty$ and $\delta < f < B_0$), we satisfy Assumption 2 of Hansen (2008). Setting $q = \infty$, $s = \infty$ in (16) and (17) of Hansen (2008) we get $\theta = \frac{\gamma - d_x - 3}{\gamma - d_x + 3}$. Therefore, since $h_N = O(N^{-\theta/d_x})$ we have

$$\frac{(\log\log N)^4 (\log N)^2}{N^\theta h_N^{d_x}} \to 0.$$

Having satisfied all the conditions of Theorem 11 of Hansen (2008), we have again that $\widehat{C}_N(z|x) \to \mathbb{E}[Y'|X = x]$ a.s.

Since $x \in \mathcal{X}$ was arbitrary we have convergence for $\mu_X$-a.e.$x$ a.s.

Now fix $D$ measurable. Consider a response variable $Y' = \mathbb{I}[y \in D]$. Then $\mathbb{E}[Y']$ exists by measurability and $Y'$ is bounded in $[0, 1]$. In addition, by Lemma 4, equal or lower mixing coefficients hold for $X, Y'$ as hold for $X, Y$. Therefore applying Theorem 11 of Hansen (2008) in the same manner again, $\hat{\mu}_{Y|x,N}(D)$ converges to $\mu_{Y|x}(D)$ for $\mu_X$-a.e.$x$ a.s.

Applying Lemma 7 we obtain that assumptions for Lemma 6 hold for $\mu_X$-a.e.$x$, a.s., which in turn yields the result desired. $\qquad\square$

*Proof of Theorem 19* Consider the unnormalized local linear weights, which we rewrite as:

$$\tilde{w}_{N,i}(x) = k_i(x) \left( 1 - \sum_{j=1}^n k_j(x)(x^j - x)^T \Xi(x)^{-1}(x^i - x) \right)$$

$$= K\left( \frac{x^i - x}{h_N} \right) \left( 1 - h_N \hat{\Phi}(x)^T \hat{\Psi}(x)^{-1} \left( \frac{x^i - x}{h_N} \right) \right),$$

$$\text{where} \quad \hat{\Phi}(x) = \frac{1}{Nh_N^{d+1}} \sum_{j=1}^n K\left( \frac{x^j - x}{h_N} \right) \left( \frac{x^j - x}{h_N} \right),$$

$$\hat{\Psi}(x) = \frac{1}{Nh_N^d} \sum_{j=1}^n K\left( \frac{x^j - x}{h_N} \right) \left( \frac{x^j - x}{h_N} \right) \left( \frac{x^j - x}{h_N} \right)^T.$$

We will show that $\tilde{w}_{N,i}(x) \geq 0$ eventually as $N$ grows for $\mu_X$-a.e.$x$, a.s. Then we will have that weights (16) are equal to weights (15) eventually and Theorem 19 applies. Let $\Sigma = \int K(u)uu^T du$, $a_n^* = \left( \frac{\log N}{Nh_N^D} \right)^{1/2} + h_N^2$, and $f_X$ denote the density of $X$. Using the properties of the kernel (symmetric and zero outside the unit ball) and the differentiability of $f_X$ (series expandable), to show that

$$\mathbb{E}\left[ \hat{\Phi}(x) \right] = \frac{1}{h_N^{d+1}} \int K\left( \frac{x' - x}{h_N} \right) \left( \frac{x' - x}{h_N} \right) f_X(u) dx'$$

$$= \frac{1}{h_N} \int K(u) u f_X(x + h_N u) du$$

$$= \frac{1}{h_N} \int K(u) u (f_X(x) + h_N \nabla f_X(x)^T u + O(h_N^2 \|u\|^2)) du$$

$$= \Sigma \nabla f_X(x) + O(h_N),$$

$$\mathbb{E}\left[ \hat{\Psi}(x) \right] = \frac{1}{h_N^d} \int K\left( \frac{x' - x}{h_N} \right) \left( \frac{x' - x}{h_N} \right) \left( \frac{x' - x}{h_N} \right)^T f_X(u) dx'$$

$$= \int K(u) uu^T f_X(x + h_N u) du$$

$$= \int K(u) uu^T (f_X(x) + h_N \nabla f_X(x)^T u + O(h_N^2 \|u\|^2)) du$$

$$= f_X(x) \Sigma + O(h_N^2).$$

By two invocations of Theorem 2 of Hansen (2008), $\hat{\Phi}(x) = \Sigma \nabla f_X(x) + O(a_n^*)$ and $\hat{\Psi}(x) = \Sigma \nabla + O(a_n^*)$ uniformly in $x$ a.s. Note that $K\left( \frac{x^i - x}{h_N} \right) = 0$ if $\left\| \frac{x^i - x}{h_N} \right\| > 1$ so to show $\tilde{w}_{N,i}(x) \geq 0$ we can restrict to $\left\|\frac{x^i - x}{h_N}\right\| \leq 1$. Therefore, we have that $\tilde{w}_{N,i}(x) = k_i(x)(1 - O(h_N))$ where $O(h_N)$ is uniformly in $x$ a.s. so that $w_{N,i}(x) \geq 0$ for all $i$ eventually for $\mu_X$-a.e.$x$, a.s. $\qquad\square$

*Proof of Theorem 20* Fix any $x$. Let us redefine

$$C(z|x) = \mathbb{E}\left[c(z; Y(z))\big|X = x\right], \qquad\qquad \widehat{C}_N(z|x) = \sum_{i=1}^N w_{N,i}(x, z_1)c(z; y^i).$$

Next, fix any $z$. Let $Y' = c(z; Y)$ and note that, by Lemma 4, the same mixing coefficients hold for $((X, Z_1), Y')$ as do for $(X, Y, Z)$. In the case of weights given by (13), applying Theorem 9 of Hansen (2008) to $((X, Z_1), Y')$ yields the following uniform convergence over the inputs to the weights $(x, z_1)$: we have that, for some $c_N \to \infty$, almost surely

$$\sup_{\|x'\| + \|z_1'\| \leq c_N} \left|\sum_{i=1}^N w_{N,i}(x', z_1')c(z; y^i) - \mathbb{E}\left[c(z; Y)\big|X = x', Z_1 = z_1'\right]\right| \to 0. \tag{31}$$

For the case of weights given by (15), applying Theorem 11 of Hansen (2008) yields the same result (31). Finally, in the case of weights given by (16), we repeat the argument in Theorem 19 verbatim but replacing $x$ by $(x, z_1)$ everywhere to arrive at the conclusion that the weights are eventually nonnegative, eventually reduce to the case of weights given by (15), and yield the same result (31). A critical observation is that (31) holds for $z_1' \neq z_1$. Now, let us restrict to the almost sure event that (31) holds simultaneously for all $z \in \mathcal{Z} \cap \mathbb{Q}^{d_z}$, of which there are countably many.

Let $\epsilon > 0$ be given. By equicontinuity, for each $z \in \mathcal{Z} \cap \mathbb{Q}^{d_z}$ there is $\delta_z > 0$ such that $|c(z; y) - c(z'; y)| \leq \epsilon/4$ for all $y, \|z - z'\| \leq \delta_z$. By density of rationals, these balls cover $\mathcal{Z}$. Since $\mathcal{Z}$ is compact, there is a finite collection $\tilde{z}^1, \ldots, \tilde{z}^k \in \mathcal{Z} \cap \mathbb{Q}^{d_z}$ such that for all $z \in \mathcal{Z}$ there is a $j$ such that $\|z - \tilde{z}^j\| \leq \delta_{\tilde{z}^j}$. Let $N_1$ be large enough so that for all $N \geq N_1$, $\|x\| + \sup_{z \in \mathcal{Z}} \|z_1\| \leq c_N$ and the left-hand-side of (31) is no more than $\epsilon/2$ for each of the finitely-many $\tilde{z}^1, \ldots, \tilde{z}^k$.

Let $N \geq N_1$. Let any $z$ be given. Let $j$ be such that $\|z - \tilde{z}^j\| \leq \delta_{\tilde{z}^j}$. Note that

$$\sum_{i=1}^N w_{N,i}(x, z_1)c(z^j; y^i) = \widehat{C}_N(z|x) + \sum_{i=1}^N w_{N,i}(x, z_1)(c(z^j; y^i) - c(z; y^i)).$$

Moreover, by Assumptions 1 and 2, we have

$$\mathbb{E}\left[c(z^j; Y)\big|X = x, Z_1 = z_1\right] = \mathbb{E}\left[c(z; Y)\big|X = x, Z_1 = z_1\right] + \mathbb{E}\left[c(z^j; Y) - c(z; Y)\big|X = x, Z_1 = z_1\right]$$

$$= \mathbb{E}\left[c(z; Y(z_1))\big|X = x\right] + \mathbb{E}\left[c(z^j; Y) - c(z; Y)\big|X = x, Z_1 = z_1\right]$$

$$= C(z|x) + \mathbb{E}\left[c(z^j; Y) - c(z; Y)\big|X = x, Z_1 = z_1\right].$$

Therefore, since $\|x\| + \|z_1\| \leq c_N$,

$$\left|\widehat{C}_N(z|x) - C(z|x)\right| = \left|\sum_{i=1}^N w_{N,i}(x, z_1)c(z; y^i) - \mathbb{E}\left[c(z; Y)\big|X = x, Z_1 = z_1\right]\right|$$

$$\leq \sum_{i=1}^N w_{N,i}(x, z_1)\left|c(z; y^i) - c(z^j; y^i)\right| + \mathbb{E}\left[\left|c(z; Y) - c(z^j; Y)\right|\big|X = x, Z_1 = z_1\right]$$

$$+ \left| \sum_{i=1}^{N} w_{N,i}(x, z_1) c(z^j; y^i) - \mathbb{E}\left[ c(z^j; Y) \big| X = x, Z_1 = z_1 \right] \right|$$

$$\leq \epsilon/4 + \epsilon/4 + \epsilon/2 = \epsilon.$$

Since $z$ and $\epsilon$ were arbitrary and $N_1$ did not depend on $z$, we have

$$\tau_N = \sup_{z \in \mathcal{Z}} \left| \widehat{C}_N(z|x) - C(z|x) \right| \to 0.$$

Therefore,

$$\delta_N = \left| \inf_{z \in \mathcal{Z}} \widehat{C}_N(z|x) - \inf_{z \in \mathcal{Z}} C(z|x) \right| \leq \sup_{z \in \mathcal{Z}} \left| \widehat{C}_N(z|x) - C(z|x) \right| = \tau_N \to 0.$$

Next, let $\hat{z}_N, \epsilon_N$ be such that $\epsilon_N \to 0$ and $\widehat{C}_N(\hat{z}_N|x) - \inf_{z \in \mathcal{Z}} \widehat{C}_N(z|x) \leq \epsilon_N$. Then,

$$0 \leq C(\hat{z}_N|x) - \inf_{z \in \mathcal{Z}} C(z|x) \leq \left| \widehat{C}_N(\hat{z}_N|x) - C(\hat{z}_N|x) \right| + \left| \widehat{C}_N(\hat{z}_N|x) - \inf_{z \in \mathcal{Z}} \widehat{C}_N(\hat{z}_N|x) \right|$$

$$+ \left| \inf_{z \in \mathcal{Z}} \widehat{C}_N(\hat{z}_N|x) - \inf_{z \in \mathcal{Z}} C(z|x) \right| \leq \tau_N + \epsilon_N + \delta_N \to 0,$$

which completes the proof. □

*Proof of Theorem 11*  By assumption of $Y$ and $V$ sharing no atoms, $\delta \overset{a.s.}{=} \tilde{\delta} \equiv \mathbb{I}\left[ Y \leq V \right]$ is observable so let us replace $\delta^i$ by $\tilde{\delta}^i$ in (24). Let

$$F(y|x) = \mathbb{E}\left[ \mathbb{I}\left[ Y > y \right] \big| X = x \right]$$

$$\hat{F}_N(y|x) = \sum_{i=1}^{N} \mathbb{I}\left[ u^i > y \right] w_{N,i}^{\text{Kaplan-Meier}}(x),$$

$$H_1(y|x) = \mathbb{E}\left[ \mathbb{I}\left[ U > y, \tilde{\delta} = 1 \right] \big| X = x \right]$$

$$\hat{H}_{1,N}(y|x) = \sum_{i=1}^{N} \mathbb{I}\left[ u^i > y, \tilde{\delta}^i = 1 \right] w_{N,i}(x),$$

$$H_2(y|x) = \mathbb{E}\left[ \mathbb{I}\left[ U > y \right] \big| X = x \right]$$

$$\hat{H}_{2,N}(y|x) = \sum_{i=1}^{N} \mathbb{I}\left[ u^i > y \right] w_{N,i}(x).$$

By assumption on conditional supports of $Y$ and $V$, $\sup\{y : F(y : x) > 0\} \leq \sup\{y : H_2(y : x) > 0\}$. By the same arguments as in Theorem 5, 6, 7, or 8, we have that, for all $y$, $\hat{H}_{1,N}(y|x) \to H_1(y|x)$, $\hat{H}_{2,N}(y|x) \to H_2(y|x)$ a.s. for $\mu_X$-a.e.x. By assumption of conditional independence and by the main result of Beran (1981), we have that, for all $y$, $F_N(y|x) \to F(y|x)$ a.s. for $\mu_X$-a.e.x. Since $\mathcal{Y}$ is a separable space we can bring the "for all $y$" inside the statement, i.e., we have weak convergence (c.f. Theorem 11.4.1 of Dudley (2002)): $\hat{\mu}_{Y|x,N} \to \mu_{Y|x}$ a.s. for $\mu_X$-a.e.x where $\hat{\mu}_{Y|x,N}$ is based on weights $w_{N,i}^{\text{Kaplan-Meier}}(x)$. Since costs are bounded, the portmanteau lemma (see Theorem 2.1 of Billingsley (1999)) gives that for each $z \in \mathcal{Z}$, $\widehat{C}_N(z|x) \to \mathbb{E}[c(z; Y)|X = x]$ where $\widehat{C}_N(z|x)$ is based on weights $w_{N,i}^{\text{Kaplan-Meier}}(x)$. Applying Lemma 7 we obtain that assumptions for Lemma 6 hold for $\mu_X$-a.e.x., a.s., which in turn yields the result desired. □

# 10. Extensions of Out-of-Sample Guarantees to Mixing Processes and Proofs

We can also extend the results of Section 8.2 to mixing processes. Combining and restating the main results of Bartlett and Mendelson (2003) (for iid) and Mohri and Rostamizadeh (2008) (for mixing), we can restate Theorem 14 as follows

THEOREM 21. *Consider a class $\mathcal{G}$ of functions $\mathcal{U} \to \mathbb{R}$ that are bounded: $|g(u)| \leq \overline{g} \ \forall g \in \mathcal{G}, \ u \in \mathcal{U}$.*
*Consider a sample $S_n = (u^1, \ldots, u^N)$ of some random variable $T \in \mathcal{T}$. Fix $\delta > 0$. If $S_N$ is generated by IID sampling, let $\delta' = \delta'' = \delta$ and $\nu = N$. If $S_N$ comes from a $\beta$-mixing process, fix some $t, \nu$ such that $2t\nu = N$, let $\delta' = \delta/2 - (\nu - 1)\beta(t)$ and $\delta'' = \delta/2 - 2(\nu - 1)\beta(t)$. Then (only for $\delta' > 0$ or $\delta'' > 0$ where they appear), we have that with probability $1 - \delta$,*

$$\mathbb{E}\left[g(T)\right] \leq \frac{1}{N} \sum_{i=1}^{N} g(u^i) + \overline{g}\sqrt{\log(1/\delta')/2\nu} + \mathfrak{R}_\nu(\mathcal{G}) \qquad \forall g \in \mathcal{G}, \tag{32}$$

*and that, again, with probability $1 - \delta$,*

$$\mathbb{E}\left[g(T)\right] \leq \frac{1}{N} \sum_{i=1}^{N} g(u^i) + 3\overline{g}\sqrt{\log(2/\delta')/2\nu} + \widehat{\mathfrak{R}}_\nu(\mathcal{G}) \qquad \forall g \in \mathcal{G}. \tag{33}$$

Replacing this result in the proof of Theorem 13 extends it to the case of data generated by a mixing process.

# 11. Proofs of Tractability Results

*Proof of Theorem 2* Let $I = \{i : w_{N,i}(x) > 0\}$, $w = (w_{N,i}(x))_{i \in I}$. Rewrite (3) as $\min w^T \theta$ over $(z, \theta) \in \mathbb{R}^{d \times n_0}$ subject to $z \in \mathcal{Z}$ and $\theta_i \geq c(z; y^i) \ \forall i \in I$. Weak optimization of a linear objective over a closed convex body is reducible to weak separation via the ellipsoid algorithm (see Grotschel et al. (1993)). A weak separation oracle for $\mathcal{Z}$ is assumed given. To separate over the $i^{\text{th}}$ cost constraint at fixed $z', \theta'_i$ call the evaluation oracle to check violation and if violated call the subgradient oracle to get $s \in \partial_z c(z'; y^i)$ with $||s||_\infty \leq 1$ and produce the cut $\theta_i \geq c(z'; y^i) + s^T(z - z')$. □

*Proof of Theorem 3* Solve (21) for each of $z_{11}, \ldots, z_{1b}$ and take the minimum. In each case, we have a problem that resembles (3) and we may use an argument similar to Theorem 2 to prove its tractability. □

*Proof of Theorem 4* Let $R = \inf_{z'_1 \in \mathcal{Z}_1} \sup_{z_1 \in \mathcal{Z}_1} ||z_1 - z'_1||$. Then no more than $b = (3RL/\epsilon)^{d_{z_1}}$ balls of radius $\epsilon/L$ are needed in order to cover $\mathcal{Z}_1$. Let their centers be denoted $z_{11}, \ldots, z_{1b}$ and apply Theorem 3. □

*Proof of Theorem 12* In the case of (25), $z(x^i) = W x^i$. By computing the norm of $W$ we have a trivial weak membership algorithm for the norm constraint and hence by Theorems 4.3.2 and 4.4.4 of Grotschel et al. (1993) we have a weak separation algorithm. By adding affine constraints $\zeta_{ij} = z_j(x^i)$, all that is left is to separate over are constraints of the form $\theta_i \geq c(\zeta_i; y^i)$, which can be done as in the proof of Theorem 2. □

## 12.   Proof of Theorem 1

*Proof*   Under Assumptions 1 and 2, the objective of problem (19) can be rewritten as

$$\mathbb{E}\left[c(z; Y(z))\big|X = x\right] = \mathbb{E}\left[c(z; Y(z_1))\big|X = x\right] \qquad \text{(By Assumption 1)}$$

$$= \mathbb{E}\left[c(z; Y(z_1))\big|X = x, Z_1 = z_1\right] \qquad \text{(By Assumption 2)}$$

$$= \mathbb{E}\left[c(z; Y(Z_1))\big|X = x, Z_1 = z_1\right] \qquad \text{(By conditioning)}$$

$$= \mathbb{E}\left[c(z; Y)\big|X = x, Z_1 = z_1\right] \qquad \text{(By definition of $Y$)}$$

which is the objective of problem (20).                                    □

## 13.   Omitted Details from Sections 1.1 and 3.2

### 13.1.   Shipment Planning Example

In our shipment planning example, we consider stocking $d_z = 4$ warehouses to serve $d_y = 12$ locations. We take locations spaced evenly on the 2-dimensional unit circle and warehouses spaced evenly on the circle of radius 0.85. The resulting network and its associated distance matrix are shown in Figure 11. We suppose shipping costs from warehouse $i$ to location $j$ are $c_{ij} = \$10 D_{ij}$ and that production costs are \$5 per unit when done in advance and \$100 per unit when done last minute.

We consider observing $d_x = 3$ demand-predictive features $X$ that, instead of iid, evolve as a 3-dimensioanl ARMA(2,2) process:

$$X(t) - \Phi_1 X(t-1) - \Phi_2 X(t-2) = U(t) + \Theta_1 U(t-1) + \Theta_2 U(t-2)$$

where $U \sim \mathcal{N}(0, \Sigma_U)$ are innovations and

$$(\Sigma_U)_{ij} = \left(\mathbb{I}\left[i = j\right]\frac{8}{7} - (-1)^{i+j}\frac{1}{7}\right)0.05,$$

$$\Phi_1 = \begin{pmatrix} 0.5 & -0.9 & 0 \\ 1.1 & -0.7 & 0 \\ 0 & 0 & 0.5 \end{pmatrix}, \quad \Phi_2 = \begin{pmatrix} 0. & -0.5 & 0 \\ -0.5 & 0 & 0 \\ 0 & 0 & 0 \end{pmatrix},$$

$$\Theta_1 = \begin{pmatrix} 0.4 & 0.8 & 0 \\ -1.1 & -0.3 & 0 \\ 0 & 0 & 0 \end{pmatrix}, \quad \Theta_2 = \begin{pmatrix} 0 & -0.8 & 0 \\ -1.1 & 0 & 0 \\ 0 & 0 & 0 \end{pmatrix}.$$

We suppose that demands are generated according to a factor model

$$Y_i = \max\{0, \, A_i^T \left(X + \delta_i/4\right) + \left(B_i^T X\right)\epsilon_i\}$$

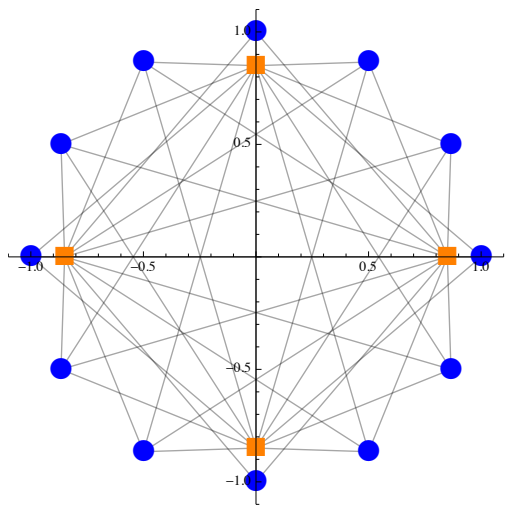

(a) The network of warehouses (orange) and locations (blue).

$$D^T = \begin{pmatrix} 0.15 & 1.3124 & 1.85 & 1.3124 \\ 0.50026 & 0.93408 & 1.7874 & 1.6039 \\ 0.93408 & 0.50026 & 1.6039 & 1.7874 \\ 1.3124 & 0.15 & 1.3124 & 1.85 \\ 1.6039 & 0.50026 & 0.93408 & 1.7874 \\ 1.7874 & 0.93408 & 0.50026 & 1.6039 \\ 1.85 & 1.3124 & 0.15 & 1.3124 \\ 1.7874 & 1.6039 & 0.50026 & 0.93408 \\ 1.6039 & 1.7874 & 0.93408 & 0.50026 \\ 1.3124 & 1.85 & 1.3124 & 0.15 \\ 0.93408 & 1.7874 & 1.6039 & 0.50026 \\ 0.50026 & 1.6039 & 1.7874 & 0.93408 \end{pmatrix}$$

(b) The distance matrix.

**Figure 11     Network data for shipment planning example.**

where $A_i$ is the mean-dependence of the $i^{\text{th}}$ demand on these factors with some idiosyncratic noise, $B_i$ the variance-dependence, and $\epsilon_i$ and $\delta_i$ are independent standard Gaussian idiosyncratic contributions. For $A$ and $B$ we use

$$A = 2.5 \times \begin{pmatrix} 0.8 & 0.1 & 0.1 \\ 0.1 & 0.8 & 0.1 \\ 0.1 & 0.1 & 0.8 \\ 0.8 & 0.1 & 0.1 \\ 0.1 & 0.8 & 0.1 \\ 0.1 & 0.1 & 0.8 \\ 0.8 & 0.1 & 0.1 \\ 0.1 & 0.8 & 0.1 \\ 0.1 & 0.1 & 0.8 \\ 0.8 & 0.1 & 0.1 \\ 0.1 & 0.8 & 0.1 \\ 0.1 & 0.1 & 0.8 \end{pmatrix}, \quad B = 7.5 \times \begin{pmatrix} 0 & -1 & -1 \\ -1 & 0 & -1 \\ -1 & -1 & 0 \\ 0 & -1 & 1 \\ -1 & 0 & 1 \\ -1 & 1 & 0 \\ 0 & 1 & -1 \\ 1 & 0 & -1 \\ 1 & -1 & 0 \\ 0 & 1 & 1 \\ 1 & 0 & 1 \\ 1 & 1 & 0 \end{pmatrix}.$$

## 13.2.    Effect of Additional Dimensions with Diminishing Predictiveness

In Section 1.1, we presented an experiment to demonstrate the effect of increasing dimension on various predictive prescriptions. In the experiment, added dimensions were noise that was a priori not distinguishable from the three main features. We can consider an alternative set up to the experiment to investigate the effect of increasing dimension when added dimensions provide additional, marginal increase in the predictiveness of $Y$. Toward that end, for each $L \in \mathbb{N}$, we consider $3L$ auxiliary variables consisting of $L$ copies of the original variables in the example where the $\ell^{\text{th}}$ copy is transformed by adding standard normal noise to the 3 variables multiplied by $(1/2)^{\ell}$. Thus, each additional copy can be used to better pin down the original variables but is more noisy than the last copy. As in Section 1.1 we fix $N = 2^{14}$. For each $L = 1, 2, 4, \ldots, 128$, we rerun the example with these alternative variables and plot the results in Figure 12.

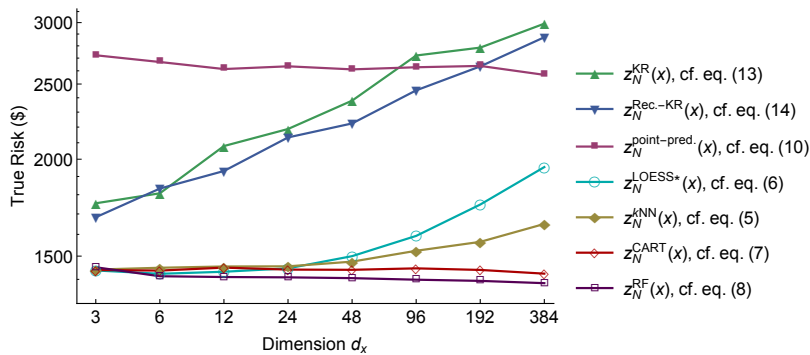

**Figure 12**      **Results of the experiment in Section 13.2.**

There are a few things to note about the results. First, the performance of the even best predictive prescriptions for $d_x = 3$ suffers due to the added noise of the first copy (compare to $d_x = 3$ in Figure 2b). Second, the performance of both the point-prediction-driven decision (using RF) and the predictive prescriptions based on CART and RF improves slowly as dimension grows and these methods use the marginal amounts of additional predictive power in the data. This shows a difference to the setting in Section 1.1, where these performed the same or very slightly worse as dimension grew. Of course, these predictive prescriptions we develop do much better than the point-prediction-driven decision, which only beats the worst ones (KR and Rec.-KR) in very high dimensions. Third, the performance of the predictive prescriptions based on KR, Rec.-KR, LOESS*, and $k$NN all have performance that deteriorates with dimension due to the curse of dimensionality and these methods' inability to learn which features are more important than others, but the deterioration is milder than in Section 1.1 and depends on the balance between the growth of dimension and the additional predictiveness offered.

## 13.3. Shipment Planning with Pricing Example

In our shipment planning with pricing example from Section 3.2, we used the same parameters as in the above shipment planning example, except that we set

$$Y_i(z_1) = \frac{100}{1 + e^{\frac{z_1 - 50}{100}}} \max\{0,\, A_i^T(X + \delta_i/4) + \left(B_i^T X\right) \epsilon_i\},$$

and we simulate the historical price data log-normally as

$$\log(Z_1) \sim \mathcal{N}(\|X\|_1 /5,\, \left|400 + e^T X\right|),$$

where $e$ is the vector of all ones.

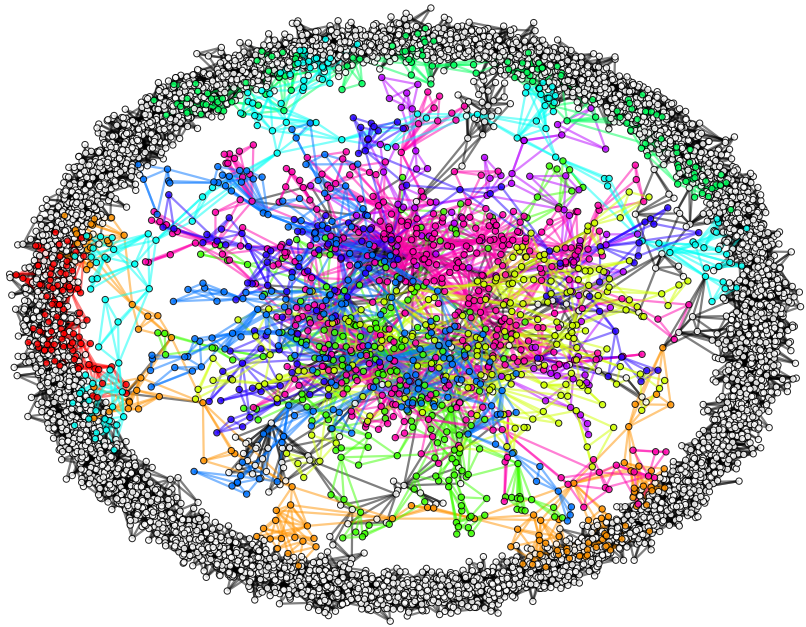

**Figure 13** **The graph of actors, connected via common movies where both are first-billed. Colored nodes correspond to the 10 largest communities of actors. Colored edges correspond to intra-community edges.**

### 13.4. Varying the Determination of $Y$

To vary the determination of $Y$ as in Section 5 we do as follows. We let $X' \in \mathbb{R}^{d_x}$ be normally distributed with 0 mean and covariance matrix equal to the covariance of $X$, $\Sigma_X$. We then introduce a new parameter $\kappa \in [0,2]$, let $\kappa' = \max\{0, 1 - \kappa\}$ and $\kappa'' = \min\{1, 2 - \kappa\}$ and redefine

$$Y_i = \max\{0, A_i^T\left(\sqrt{\kappa}X + \sqrt{\kappa'}X' + \sqrt{\kappa''}\delta_i/4\right) + \sqrt{\kappa''}\sqrt{\kappa}\left(B_i^T X\right)\epsilon_i + \sqrt{\kappa''}\sqrt{\kappa'}(V\Sigma_X V^T)_i^{T/2}\epsilon\}.$$

The original example corresponds to $\kappa = 1$, whereas $\kappa = 0$ corresponds to independence between $X$ and $Y$ and $\kappa = 2$ corresponds to $Y$ being measurable with respect to (a function of) $X$.

## 14. Constructing Auxiliary Data Features from IMDb and RT data

For some information harvested from IMDb and RT, the corresponding numeric feature is straightforward (e.g. number of awards). For other pieces of information, some distillation is necessary. For genre, we create an indicator vector. For MPAA rating, we create a single ordinal (from 1 for G to 5 for NC-17). For plot, we measure the cosine-similarity between plots,

$$\text{similarity}(P_1, P_2) = \frac{p_1^T p_2}{||p_1||\,||p_2||},$$

where $p_{ki}$ denotes the number of times word $i$ appears in plot text $P_k$ and $i$ indexes the collection of unique words appearing in plots $P_1$, $P_2$ ignoring certain generic words like "the". and use this as a distance measure to hierarchically cluster the plots using Ward's method (cf. Ward (1963)). This captures common themes in titles. We construct 12 clusters based solely on historical data and, for new data, include a feature vector of median cosine similarity to each of the clusters. For actors, we create a graph with titles as nodes and with edges between titles that share actors, weighted by the number of actors shared. We use the method of Blondel et al. (2008) to find communities of titles and create an actor-counter vector for memberships in the 10 largest communities (see Figure 13). This approach is motivated by the existence of such actor groups as the "Rat Pack" (Humphery Bogart and friends), "Brat Pack" (Molly Ringwald and friends), and "Frat Pack" (Owen Wilson and friends) that often co-star in titles with a similar theme, style, and target audience.